\documentclass[10pt,twocolumn,letterpaper]{article}
\usepackage{cvpr}
\usepackage{times}
\usepackage{epsfig}
\usepackage{graphicx}
\usepackage{amsmath}
\usepackage{amssymb}

\usepackage{url}
\usepackage{adjustbox}
\usepackage{array}
\usepackage{multirow}
\usepackage{threeparttable}
\usepackage{soul}
\newcommand{\tabincell}[2]{\begin{tabular}{@{}#1@{}}#2\end{tabular}}  
\newcommand\ChangeRT[1]{\noalign{\hrule height #1}}
\usepackage[nomargin, inline,marginclue,draft]{fixme}
\usepackage{xcolor}

\cvprfinalcopy 

\def\cvprPaperID{5} 

\ifcvprfinal\fi
\begin{document}

\definecolor{Jimmy}{rgb}{0.5, 0.2, 0.7}
\definecolor{YS}{rgb}{0.0, 0.67, 0.28}
\definecolor{Ulia}{rgb}{0.7, 0.45, 0.2}
\definecolor{YM}{rgb}{0.8, 0.3, 0.6}
\definecolor{HKKUO}{rgb}{0.36, 0.54, 0.66}
\definecolor{WT}{rgb}{0, 0, 0.7}
\definecolor{roy}{rgb}{0.45, 0.31, 0.59}
\definecolor{ISSUE}{rgb}{1.0, 0, 0}
\definecolor{REVISE}{rgb}{0.8, 0.8, 0.8}
\definecolor{STEP2}{rgb}{0.44, 0.19, 0.63}
\definecolor{STEP3}{rgb}{0.77, 0.35, 0.07}
\definecolor{STEP4}{rgb}{1.0, 0.75, 0.0}
\definecolor{FINALv1}{rgb}{1.0, 0, 0}
\definecolor{FINALv2}{rgb}{0, 0, 1.0}

\title{Deploying Image Deblurring across Mobile Devices: A Perspective of Quality and Latency}

\author{Cheng-Ming Chiang\thanks{Email: jimmy.chiang@mediatek.com} \and Yu Tseng \and Yu-Syuan Xu \and Hsien-Kai Kuo \and Yi-Min Tsai \and Guan-Yu Chen \and Koan-Sin Tan \and Wei-Ting Wang \and Yu-Chieh Lin \and Shou-Yao Roy Tseng \and Wei-Shiang Lin \and Chia-Lin Yu \and BY Shen \and Kloze Kao \and Chia-Ming Cheng \and Hung-Jen Chen \\
MediaTek Inc., Hsinchu, Taiwan\\}


\maketitle
\thispagestyle{empty}

\begin{abstract}
Recently, image enhancement and restoration have become important applications on mobile devices, such as super-resolution and image deblurring.
However, most state-of-the-art networks present extremely high computational complexity.
This makes them difficult to be deployed on mobile devices with acceptable latency.
Moreover, when deploying to different mobile devices, there is a large latency variation due to the difference and limitation of deep learning accelerators on mobile devices.
In this paper, we conduct a search of portable network architectures for better quality-latency trade-off across mobile devices.
We further present the effectiveness of widely used network optimizations for image deblurring task.
This paper provides comprehensive experiments and comparisons to uncover the in-depth analysis for both latency and image quality.
Through all the above works, we demonstrate the successful deployment of image deblurring application on mobile devices with the acceleration of deep learning accelerators.
To the best of our knowledge, this is the first paper that addresses all the deployment issues of image deblurring task across mobile devices.
This paper provides practical deployment-guidelines, and is adopted by the championship-winning team in NTIRE 2020 Image Deblurring Challenge on Smartphone Track.
\end{abstract}

\section{Introduction}

Deep learning based networks have achieved great successes in image enhancement and restoration tasks \cite{srgan, rdn, sid, deblurganv2, pynet}.
Among these applications, image deblurring have become one of the most important camera features for mobile devices \cite{mobiledeblur}.
Due to the large input resolution and the characteristic of pixel-to-pixel mapping nature, these contemporary networks demand extremely high complexity and memory footprint. 
This makes deploying an image deblurring task on mobile devices a great challenge.

In recent years, deep learning communities have noticed the gap between network design and its deployment on mobile devices.
Image enhancement and restoration on smartphone contests have been held to shed a light upon this problem \cite{pirmsr, ntiredeblur}. 
Meanwhile, deep learning accelerators are also widely adopted in most of mobile devices \cite{s855, dimensity1000, kirin990, exynos990, iphone11}.
Following the trend, some benchmark suites are proposed to evaluate the performance of these mobile devices \cite{aibenchmark, mlperf, ludashi, antutu}.
To alleviate the burden of network deployment, some papers propose light-weight network architectures to reduce the complexity \cite{idn, carn, falsr, mtlu, mobilenet}.
Another idea is network optimization, which targets on arbitrary network architectures. 
Among the existing technologies, quantization \cite{tfquant} and pruning \cite{deepcompression} are two of the most popular techniques to optimize network performance.

However, the applicability of these optimization techniques with respect to image deblurring task is rarely discussed.
Furthermore, the performance of a network is highly affected by the hardware limitations and preferences.
Therefore, network portability is another key factor to deploy across a set of mobile devices.
Last but not the least, the existing benckmarking efforts lack of a realistic setting to well reflect the practical use-cases, \eg, 720p High-Definition (HD) input resolution ($1280\times720$).



In this paper, we compare both quality and latency index of different image deblurring networks across mobile devices.
Practical settings are adopted to reflect real user scenarios.
Our contributions are summarized as below:
\begin{itemize}
\item {\bf Portable Network Architectures.}
We conduct a search of portable network architectures for better quality-latency trade-off across mobile devices.
This also includes a set of practical application settings to better reflect real user scenarios.

\item {\bf Network Optimization.}
For image deblurring task, we further present the effectiveness of popular network optimizations, quantization and pruning.
We demonstrate that there exist noticeable quality drops with 8-bit quantization-aware training.
With 16-bit post-training quantization, it is capable of achieving the same quality level as floating-point network.

\item {\bf Quality and Latency across Mobile Devices.}
In terms of image quality and latency, we evaluate various image deblurring networks across mobile devices.
Our paper demonstrates the success deployment of image deblurring application on three mobile devices (with deep learning accelerators).

\end{itemize}

In Section~\ref{sec:related}, we describe the related work for this paper.
We will introduce detailed flow of deploying image deblurring on mobile devices in Section~\ref{sec:method}.
In Section~\ref{sec:result}, we show detailed analysis from the aspect of quality and latency.
The conclusion and future work are summarized in Section~\ref{sec:conclusion}.

\section{Related Work}
\label{sec:related}

\subsection{Image Enhancement and Restoration}

In recent works, most of the image enhancement and restoration methods share common network architectures.
U-Net \cite{unet} architecture, which is also known as encoder-decoder structure, is widely used in many image enhancement and restoration tasks \cite{scalerecurrent, dynamicscenedeblur, aliveblur, noise2noise, megviinr, sid}.
In image denosing, Gu \etal~\cite{sgn} propose a top-down architecture, Self-Guided Network (SGN), to better exploit multi-scale information in images.
In super resolution, there are also quite a few representative network architectures, such as EDSR \cite{edsr}, RDN \cite{rdn} and DBPN \cite{dbpn}.
Most of these architectures keep the same scale across all operations except the last one, which is responsible for up-sampling.
In image deblurring, besides U-Net architecture, deformable convolution and self-attention module are proposed to model spatially-varying deblurring process in \cite{deformabledeblur}. Recently, Kupyn \etal~\cite{deblurganv2} use FPN architecture \cite{fpn} in image deblurring with Generative Adversarial Network (GAN) based training methodology.
To alleviate the computational complexity of deployment, several light-weight architectures are proposed recently \cite{idn, carn, falsr, mtlu, mobilenet}.


\subsection{Network Optimizations}

\paragraph{Quantization.} 
Network quantization is one of the most effective methods for deploying networks on mobile devices. 
Typically, quantization enables efficient integer arithmetics by translating weights and activations of a network into fixed-point (\eg, 8-bit integer) representation. 
Quantization-aware training and post-training quantization are two well-known techniques supported by TensorFlow \cite{tfquant}.
Post-training quantization estimates value ranges for both weights and activations through forward pass of training data while quantization-aware training performs such estimation in both forward and backward pass.
In recent works, both techniques demonstrate promising results on image perception tasks \cite{tfquant, resnet, inceptionv3, mobilenet, tfdetection, tfdeeplabquant}. 
To the best of our knowledge, there are limited works \cite{srquant, binarysr} applying quantization on image enhancement problems. 
To better understand the effectiveness of quantization on image deblurring, this work applies the most widely used quantization techniques and conducts a comprehensive evaluation.

\paragraph{Pruning.}  
Network pruning is another widely used optimization for deploying networks on mobile devices. 
There are two approaches of pruning, unstructured pruning \cite{deepcompression, mitprune} and structured pruning \cite{morphnet, oicsr, gatedecorator}. Unstructured pruning makes the weights of a network sparse instead of changing the network architecture.
Structured pruning reduces the number of channels in the network and thus improves latency on general devices. 
Most of the works focus on image classification or segmentation \cite{deepcompression, mitprune, morphnet, oicsr, gatedecorator}. 
Wang \etal~\cite{mtkprune} propose architecture-aware pruning to reduce MAC\footnote{MAC is known as multiply-accumulate. A MAC is roughly two floating-point operations (FLOPs), used in some other papers.} and memory bandwidth in super resolution \cite{edsr} and low-light enhancement \cite{sid}.
In this work, we apply pruning techniques in similar ways and show the effectiveness on image deblurring.

\subsection{Benchmark Suites \& Challenges}

\paragraph{Benchmark Suites.}
\emph{AI Benchmark}, is a comprehensive benchmark suite for mobile devices by Andrey \etal~\cite{aibenchmark}, which evaluates both latency and accuracy among various tasks.
In \emph{AI Benchmark}, the resolution for input images are ranging from $84\times84$ to $512\times512$ (except semantic segmentation which is not the focus of this paper). 
However, contemporary use-cases of image enhancement typically need larger input resolution, for example, 720p HD or even higher resolution.
\emph{MLPerf inference benchmark} \cite{mlperf} is one of the largest benchmark community contributed from both academic and industry. However, image enhancement and restoration tasks are not included for its benchmarking. 
\emph{AIMark} \cite{ludashi} and \emph{Antutu AI Benchmark} \cite{antutu} are another two benchmark suites targeting mobile devices.
Among these two benchmark suites, platform providers are asked to deploy test applications by using proprietary formats and frameworks.
Such benchmarking policy is quiet different from \emph{AI Benchmark} which adopts a unified framework, Android Neural Networks API (NNAPI) \cite{nnapi}.

\begin{figure*}[t]
\begin{minipage}[b]{1\linewidth}
  \centering
  \centerline{\includegraphics[width=18cm]
  {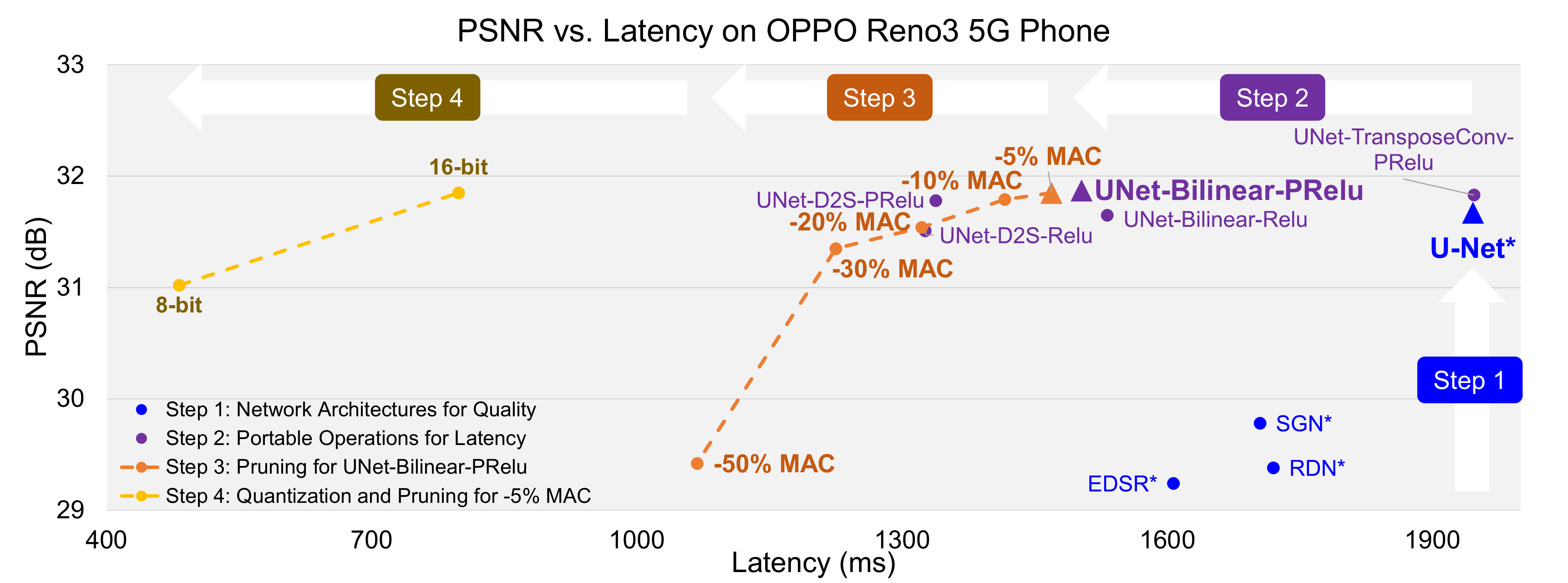}}
  \medskip
\end{minipage}
\vspace{-15pt}
\caption{The evolution of trade-off between PSNR and latency on OPPO Reno3 5G.}

\label{fig:explorationflow}
\vspace{-5pt}
\line(1,0){250}\\
\footnotesize (*) means the architecture is slightly different from the original paper.\\
\footnotesize U-Net*: UNet-TransposeConv-Relu.\\
\footnotesize D2S: abbreviation of DepthToSpace. Bilinear: abbreviation of ResizeBilinear.
\end{figure*}

\paragraph{Challenges.} 
\emph{PIRM 2018 challenge on perceptual image enhancement on smartphone} \cite{pirmsr} is the first image enhancement challenge that evaluates latency on mobile devices.
Razer phone and Huawei P20 are used as target devices \cite{razarp20}, which have their latest generations with higher computation capacity. 
\emph{NTIRE 2020 image deblurring challenge on smartphone} \cite{ntiredeblur} adopts Google Pixel 4 as its target device. 
However, the evaluation of latency is conducted on $256\times256$ input resolution, which is far from enough to reflect a real use-case, say HD 720p ($1280\times720$). 

In this paper, we apply a more realistic setting for image deblurring application and deploy it across a set of mobile devices.
Differentiation includes 
(1) We use HD 720p resolution ($1280\times720$) to show the capability of mobile devices for real use-cases. 
(2) We evaluate both quantitative (PSNR\footnote{Peak Signal to Noise Ratio}) and qualitative (visual) results to better justify the quality measurement.
(3) To create a fair comparison across mobile devices, we adopt the unified NNAPI framework \cite{nnapi} for all the evaluations, including both quality and latency.

\section{Deploying Image Deblurring across Mobile Devices}
\label{sec:method}
In this paper, we take image deblurring as an example for the mobile deployment.
We first introduce the problem definition of blind image deblurring.
Second, we elaborate the searching procedure of portable network architecture and its interplay between latency portability and PSNR quality.
Then, we describe the optimization techniques to further improve performance on mobile devices.
Finally, softwares and hardwares for deploying the networks are introduced.

\subsection{Image Deblurring}
In this paper, we adopt the same problem formulation, blind image deblurring task, as used in \emph{NTIRE 2020 image deblurring challenge on smartphone} \cite{ntiredeblur}.


\paragraph{Dataset.}
For the training dataset, we use REDS \cite{reds} image deblurring dataset which is also used in image deblurring challenges of NTIRE 2020 \cite{ntiredeblur}. In REDS dataset, there are 300 videos divided into 240 sequences for training, 30 sequences for validation, and 30 sequences for testing. Each sequence contains 100 frames of $1280\times720$ resolution. For each frame, blurry image and sharp image are given as a pair. In this paper, we treat each frame as independent and conduct all the experiments with this setting.

\begin{table*}
\begin{center}
    \centering
    \caption{Hardware specification and AI-Scores \cite{aibenchmarkweb} of the mobile devices.}
    \label{tab:hwspec}
    \begin{tabular}{| c | c | c | c |}
        \hline
    	& \textbf{Huawei Mate30 Pro 5G} & \textbf{OPPO Reno3 5G} & \textbf{Google Pixel 4} \\
    	\hline\hline
    	Chipset & HiSilicon Kirin 990 5G \cite{kirin990} & MediaTek Dimensity 1000L \cite{dimensity1000} &
    	Qualcomm Snapdragon 855 \cite{s855} \\
    	\hline
    	CPU & \tabincell{c}{2 Cortex-A76, 2.86 GHz \\ 2 Cortex-A76, 2.36 GHz \\ 4 Cortex-A55, 1.95 GHz} &
    	\tabincell{c}{4 Cortex-A77, 2.20 GHz \\ 4 Cortex-A55, 2.00 GHz} &
    	\tabincell{c}{1 Kryo 485 Gold Prime, 2.84 GHz \\ 3 Kryo 485 Gold, 2.42 GHz \\ 4 Kryo 485 Silver, 1.80 GHz} \\
    	\hline
    	GPU & Mali-G76 & Mali-G77 & Adreno 640 \\
    	\hline
    	AI Engine & \tabincell{c}{2 Big-Core DaVinci NPU \\ 1 Tiny-Core DaVinci NPU} & \tabincell{c}{APU 3.0 (2 Big Cores,\\3 Small Cores, 1 Tiny Core)} & AIE CPU, AIE GPU, AIE DSP \\
    	\hline
    	\tabincell{c}{NNAPI\\Runtime} & \tabincell{c}{nnapi-reference, armnn, \\liteadaptor} & \tabincell{c}{nnapi-reference, neuron-ann} &
    	\tabincell{c}{nnapi-reference, google-edgutpu, \\qti-default, qti-dsp, qti-gpu, qti-hta}\\
    	\hline\hline
    	\tabincell{c}{AI-Score \cite{aibenchmarkweb}} & 76,206 & 58,628 & 33,289 \\
    	\hline
    \end{tabular}
\end{center}
\vspace{-15pt}
\end{table*}

\subsection{Searching Portable Network Architectures}
\label{sec:searching}
In this section, we introduce the searching of a high PSNR quality yet portable network across mobile devices.
This includes Step 1 and Step 2 shown in Figure~\ref{fig:explorationflow}.
First of all, a set of state-of-the-art network architectures is listed in Figure~\ref{fig:explorationflow}.
For fair comparison, networks are slightly adjusted to match a baseline computational complexity (refer to Section~\ref{sec:result} for more detail).
In \textcolor{blue}{Step 1}, the goal is to search for the architecture with highest quality.
Hence, the network of the highest PSNR, \textcolor{blue}{U-Net} \cite{unet}, is selected for the next step.
In \textcolor{STEP2}{Step 2}, the objective turns to increase the portability across difference mobile devices.
In accelerator hardware, optimization usually focuses on limited operations, such as convolution, pooling, activation and so on.
Therefore, a network with high quantitative or qualitative quality can have very limited portability since its operations are not optimized on another devices.
Thankful to the strong function approximating nature of neural network, it is possible to replace these operations by other semantically similar and optimized ones.
According to the above discussion, we derive a set of architectures from the previous step.
These architectures are marked as \textcolor{STEP2}{purple} in Figure~\ref{fig:explorationflow} (refer to Section~\ref{sec:result} for more detailed discussions).
To this end, one is free to choose any of the networks according to the quality or latency requirement.

\subsection{Optimizing Network for Deployment}
\label{sec:optimizing}
As discussed in Section~\ref{sec:searching}, the procedure searches for a set of portable network architectures with the best quality-latency trade-off. After that, several network optimization techniques can be applied to further boost the performance on mobile devices. The following paragraphs introduce how the widely used pruning and quantization are applied to the portable network architectures.

\vspace{-10pt}
\paragraph{Pruning.} 
We use structured pruning technique to optimize networks (\textcolor{STEP3}{Step 3} in Figure~\ref{fig:explorationflow}). A structured pruning technique similar to one mentioned in \cite{mtkprune} is used.
The numbers of channels are adaptively pruned with respect to the given MAC reduction target.
With different level of pruning targets, a set of pruned networks are generated and marked \textcolor{STEP3}{orange} as in Figure~\ref{fig:explorationflow}. Any choice is a trade-off between quality and latency.

\vspace{-10pt}
\paragraph{Quantization.} 
We apply quantization-aware training and post-training quantization techniques to demonstrate their applicability on the image deblurring task.
We extend the evaluation to both 8-bit and 16-bit quantization, which will be detailed in Section~\ref{sec:exp_opt}.
Similarly, in \textcolor{STEP4}{Step 4}, the quantized networks provide another opportunity for trade-off.

\begin{figure}[t]
\begin{minipage}[b]{1\linewidth}
  \centering
  \centerline{\includegraphics[width=8.4cm]
  {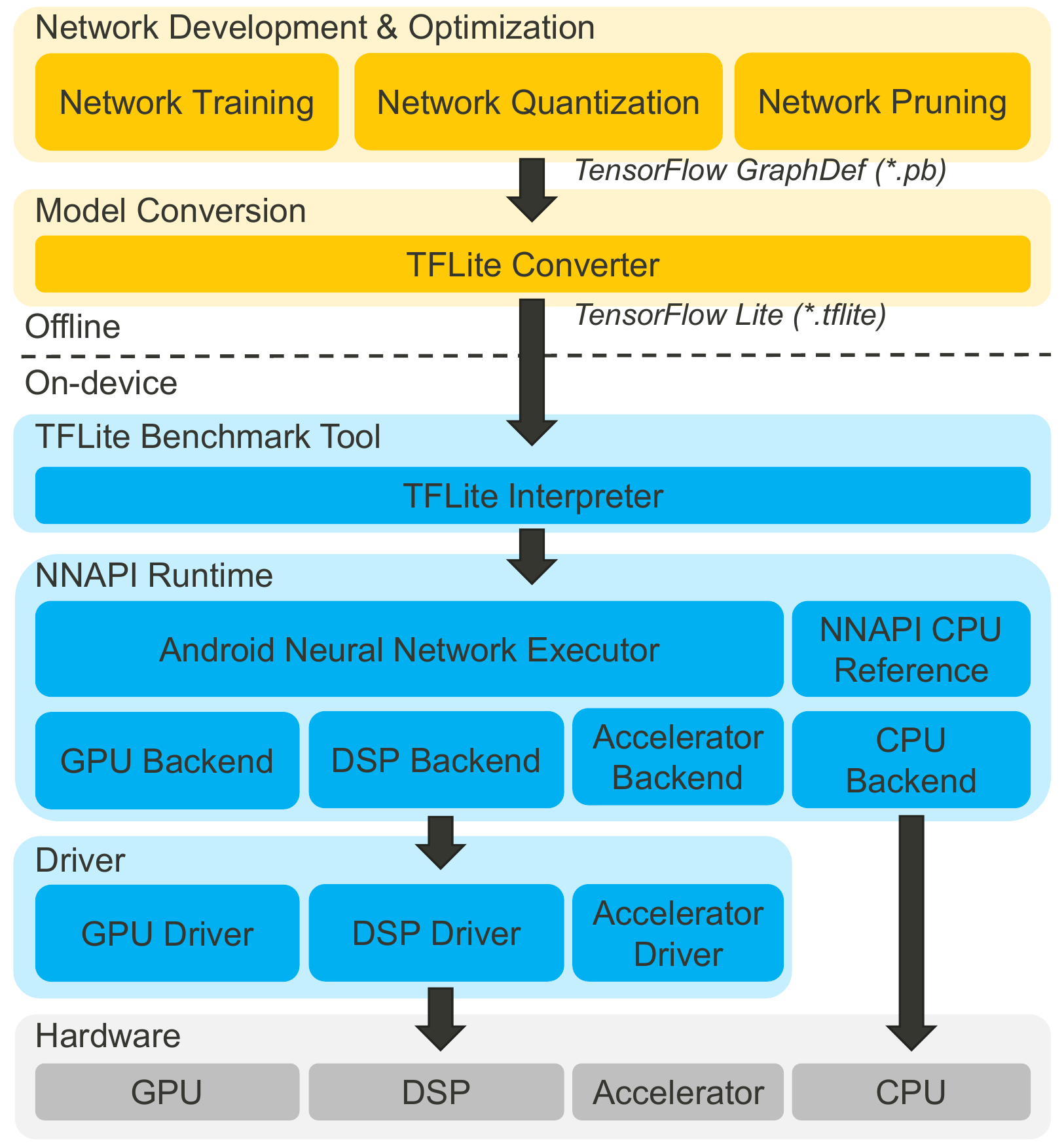}}
  \medskip
\end{minipage}
\caption{Software stack for developing networks, optimizing networks, and inference on mobile devices.}
\label{fig:swstack}
\vspace{-10pt}
\end{figure}

\subsection{Deploying Network: Softwares and Hardwares}
As shown in Figure~\ref{fig:swstack}, this paper adopts TFLite format (.tflite) and \emph{TFLite Benchmark Tool} \cite{benchmarktool} to evaluate the latency on various mobile devices. NNAPI, underneath TFLite, is a unified inference framework widely supported by various platforms \cite{kirin990, dimensity1000, s855}. For fair comparison across mobile devices as in \cite{aibenchmark}, we adopt the unified NNAPI framework to deploy the networks on mobile devices. Last, we deploy the optimized portable networks across several mobile devices and conduct the experimental analysis.

\begin{table*}
\begin{center}
    \centering
    \caption{Quality, complexity, and latency of different network architectures.}
    \label{tab:archicompare}
    \begin{threeparttable}[t]
    \begin{tabular}{ c  c  c  c  c  c }
        \ChangeRT{0.5pt}
        \hline
        \multirow{2}{*}{\tabincell{c}{Network}} & \multirow{2}{*}{\tabincell{c}{PSNR / SSIM \\ (floating-point)}} & 
        \multirow{2}{*}{\tabincell{c}{MAC \\($\times10^{9}$)}} & \multicolumn{3}{c}{Latency (ms)} \\ \cline{4-6}
        & & & Huawei Mate30 Pro & OPPO Reno3 5G & Google Pixel 4 \\ 
        \hline
        U-Net \cite{unet}*                         & \bf{31.67 / 0.899} & 238 & 54206\tnote{1}  & \bf{1946}      & 31870\tnote{2} \\
        EDSR \cite{edsr}*                          & 29.24 / 0.836      & 249 & \bf{518}        & 1607           & 3468 \\
        RDN \cite{rdn}*                            & 29.40 / 0.839      & 243 & 7367            & \bf{1720}      & Failed\tnote{3} \\
        DBPN \cite{dbpn}*                          & 31.23 / 0.886      & 242 & Failed\tnote{3} & Failed\tnote{3}& Failed\tnote{3} \\
        Inception-ResNetV2-FPN \cite{deblurganv2}* & 28.63 / 0.830      & 250 & 22164\tnote{4}  & \bf{3245}      & 20150\tnote{5} \\
        SGN \cite{sgn}*                            & 29.78 / 0.858      & 247 & \bf{931}        & 1705           & 3004\tnote{6} \\
        \ChangeRT{0.5pt}
        \hline

    \end{tabular}
    \begin{tablenotes}\scriptsize
    \item (*) means the architecture is slightly different from the original paper.
    \item[1] TRANSPOSE\_CONV\_2D operations fall back to CPU.
    \item[2] TRANSPOSE\_CONV\_2D, CONCATENATION and CONV\_2D operations after CONCATENATION fall back to CPU.
    \item[3] ERROR: NN API returned error ANEURALNETWORKS\_OP\_FAILED.
    \item[4] MUL and RESIZE\_NEAREST\_NEIGHBOR operations fall back to CPU.
    \item[5] MUL, RESIZE\_NEAREST\_NEIGHBOR, some CONV\_2D and MAP\_POOL\_2D operations fall back to CPU.
    \item[6] SPACE\_TO\_DEPTH and DEPTH\_TO\_SPACE operations fall back to CPU.
    
    \end{tablenotes}
    \end{threeparttable}
\end{center}
\vspace{-20pt}
\end{table*}

\vspace{-10pt}
\paragraph{TFLite Benchmark Tool.}
\emph{TFLite Benchmark Tool} \cite{benchmarktool} can be used to evaluate the latency of a TFLite model on both desktops and Android devices.
It provides several accelerations on mobile devices, \eg, XNNPACK delegate is optimized for floating-point inference on ARM CPU, 
GPU delegate for floating-point inference on mobile GPU, 
NNAPI delegate for both floating-point and 8-bit fixed-point inference on Android devices, and 
Hexagon delegate is optimized for 8-bit fixed-point inference on Qualcomm DSP. 
With these tools, comprehensive latency evaluation can be conducted on different mobile devices.

\vspace{-10pt}
\paragraph{Android NNAPI.} NNAPI \cite{nnapi} is designed for accelerating deep learning operations on Android devices. It provides base operators of functionality for higher-level machine learning frameworks, such as TensorFlow Lite (TFLite) and Caffe2. 
With NNAPI, platform providers can have specific acceleration of frequently used operations for both IEEE 754 16-bit floating-point and 8-bit fixed-point data type.

\paragraph{Hardware Acceleration.} 
This paper focuses on three mobile devices (with deep learning accelerators) as in \emph{AI-Score} \cite{aibenchmarkweb}, including Huawei Mate30 Pro 5G, OPPO Reno3 5G, and Google Pixel 4 \cite{phonecompare}. Table~\ref{tab:hwspec} summarizes the hardware specification of these mobile devices.
Table~\ref{tab:hwspec} also lists the details of NNAPI runtime library for each platform.

\section{Experiment Results}
\label{sec:result}
In this section, we discuss the experiment results and its implementation details. 
Section~\ref{sec:implementation} elaborates the details of dataset, training setups and evaluation methods. 
The proposed architecture search is discussed in Section~\ref{sec:exp_arch}. Section~\ref{sec:exp_ops} covers the portability discussion at operation level. 
The interplay between networks and optimization methods are discussed in Section~\ref{sec:exp_opt}. 
Last but not the least, we discuss the quality considerations in Section~\ref{sec:exp_quality}.

\subsection{Implementation Details}
\label{sec:implementation}
In this paper, we implement and train all the networks by TensorFlow.
We crop each training image in REDS dataset \cite{reds} into 15 patches, with $256\times256$ resolution for each patch.
A total of 360,000 pairs of training patches are prepared from 240 training sequences (100 frames for each).
We follow the \emph{NTIRE 2020 image deblurring challenge on smartphone} for frame selection (1 out of every 10 frames) \cite{ntiredeblur}.
As a result, a total of 300 frames in validation set are used to calculate PSNR for quality assessment.

All networks are trained for 1M steps on a single RTX-2080 Ti GPU with batch size 16, $L_1$ loss, Adam optimizer \cite{adam} and exponential decay for learning rate.
We set initial learning rate as $2\times10^{-4}$, decay rate as 0.98, and 5K decay steps for exponential decay.
All convolutional operations are initialized with Xavier initialization \cite{xavier}.

To measure the latency on mobile devices, we use \emph{TFLite Benchmark Tool} \cite{benchmarktool} with arguments $use\_nnapi=true$, $allow\_fp16=true$\footnote{To inference floating-point networks with 16-bit floating-point data type}, $num\_runs=10$, and  $num\_threads=4$.
We also use \emph{taskset}\footnote{We use "taskset f0" to specify using 4 big cores of CPU} command to reduce the variation of CPU  time between different runs.

\begin{table*}
\begin{center}
    \centering
    \caption{Quality, complexity, and latency of different up-sampling and activation operations for U-Net.}
    \label{tab:opcompare}
    \begin{threeparttable}[t]
    \begin{tabular}{ c  c  c  c  c  c }
        \ChangeRT{0.5pt}
        \hline
        \multirow{2}{*}{\tabincell{c}{Network}} & \multirow{2}{*}{\tabincell{c}{PSNR / SSIM \\ (floating-point)}} & 
        \multirow{2}{*}{\tabincell{c}{MAC \\($\times10^{9}$)}} & \multicolumn{3}{c}{Latency (ms)} \\ \cline{4-6}
        & & & Huawei Mate30 Pro & OPPO Reno3 5G & Google Pixel 4 \\ 
        \hline
        UNet-TransposeConv-Relu\tnote{\dag} & 31.67 / 0.899 & \multirow{2}{*}{238} & 54206\tnote{1} & \bf{1946} & 31870\tnote{2} \\
        UNet-TransposeConv-PRelu             & 31.83 / 0.903 &                      & 78402\tnote{1} & \bf{1947} & 32390\tnote{2} \\
        UNet-DepthToSpace-Relu               & 31.51 / 0.895 & \multirow{2}{*}{222} & \bf{805}       & 1326      & 2697\tnote{3} \\
        UNet-DepthToSpace-PRelu              & 31.78 / 0.900 &                      & \bf{908}       & 1338      & 3060\tnote{3} \\
        UNet-ResizeBilinear-Relu             & 31.65 / 0.898 & \multirow{2}{*}{256} & \bf{1184}      & 1532      & 8425\tnote{4} \\
        UNet-ResizeBilinear-PRelu            & \bf{31.87 / 0.903} &                 & \bf{1281}      & 1503      & 9770\tnote{4} \\
        \ChangeRT{0.5pt}
        \hline

    \end{tabular}
    \begin{tablenotes}\scriptsize
    \item[\dag] UNet-TransposeConv-Relu is the same as U-Net \cite{unet}* in Table~\ref{tab:archicompare}
    \item[1] TRANSPOSE\_CONV\_2D operations fall back to CPU.
    \item[2] TRANSPOSE\_CONV\_2D, CONCATENATION and CONV\_2D operations after CONCATENATION fall back to CPU.
    \item[3] DEPTH\_TO\_SPACE operations fall back to CPU.
    \item[4] CONCATENATION and CONV\_2D after CONCATENATION operations fall back to CPU.
    \end{tablenotes}
    \end{threeparttable}
\end{center}
\vspace{-5pt}
\end{table*}

\subsection{Network Architectures for Quality}
\label{sec:exp_arch}
To search across different network architectures, we compare some widely used networks in image enhancement domain,
including U-Net \cite{unet}, Inception-ResNetV2-FPN \cite{deblurganv2}, EDSR \cite{edsr}, RDN \cite{rdn}, DBPN \cite {dbpn}, and SGN \cite{sgn}.
For fair comparison, a network's operations and channels are slightly adjusted to match a baseline computational complexity, roughly $250\times10^{9}$ MAC.
We remove up-sampling in EDSR and RDN, since these operations were designed for super resolution.
Likewise, in DBPN, we remove the first up-projection unit to keep the same resolution for input and output tensors.
We exclude deformable convolution and self-attention based networks \cite{deformabledeblur} since no mobile device supports these types of operations.
Knowing that this paper focuses on deploying the architectures on mobile devices, any training methodology can also be applied to the architectures of interest, \eg, GAN-based training.
The detail training settings are summarized in Section~\ref{sec:implementation}.

Table~\ref{tab:archicompare} summarizes PSNR and MAC of the candidates of network architectures.
We also list the measured latency of these networks on all the three target mobile devices.
According to the quality index in Table~\ref{tab:archicompare}, U-Net outperforms all the counterparts by its highest PSNR.
However, an interesting finding is that, an unsupported operation by accelerator\footnote{We use "adb shell setprop debug.nn.vlog 1" to open debug option and use "adb shell logcat | grep -e findBestDeviceForEachOperation" to check whether an operation is executed on CPU or accelerator}, \eg, \emph{TRANSPOSE\_CONV\_2D}, will cause a fallback to NNAPI CPU-reference-implementation. 
This prevents the execution from being accelerated and results in unreasonable high latency as shown in Table~\ref{tab:archicompare}.

\subsection{Portable Operations for Latency}
\label{sec:exp_ops}

As discussed in Section~\ref{sec:exp_arch}, an unsupported operation across devices can result in unreasonable high latency.
The major functionality of \emph{TRANSPOSE\_CONV\_2D} operation is for up-sampling.
Hence, in order to deploy the network across all the three mobile devices, an alternative solution is to replace such operations by other operations (with similar functionality).
In this paper, we replace \emph{TRANSPOSE\_CONV\_2D} by \emph{DEPTH\_TO\_SPACE}\footnote{DEPTH\_TO\_SPACE is also known as pixel shuffle in some papers or frameworks} and \emph{RESIZE\_BILINEAR}.
The replacement is also evaluated on both \emph{RELU} and \emph{PRELU} activations to show its effectiveness.

As shown in Table~\ref{tab:opcompare}, such replacement avoids most of the cases in which a fallback to NNAPI CPU-reference-implementation happens.
Thus, a network with better trade-off between quality and latency can be conducted in this way.
 One is free to choose any of the networks according to the quality or latency.
In this paper, \emph{UNet-ResizeBilinear-PRelu} is selected for the following experiments.

\begin{table*}
\begin{center}
    \centering
    \caption{Quality, complexity, and latency of different optimization techniques.}
    \label{tab:optimizationcompare}
    \begin{threeparttable}[t]
    \begin{tabular}{ c  c  c  c  c  c  c  c }
        \ChangeRT{0.5pt}
        \hline
        \multirow{3}{*}{\tabincell{c}{Network}} & \multirow{3}{*}{\tabincell{c}{Optimization\\Type}} & \multirow{3}{*}{Setting} & \multirow{3}{*}{PSNR / SSIM} & 
        \multirow{3}{*}{\tabincell{c}{MAC\\($\times10^{9}$)}} & \multicolumn{3}{c}{Latency (ms)} \\ \cline{6-8}
        & & & & & \tabincell{c}{Huawei\\Mate30 Pro} & \tabincell{c}{OPPO\\Reno3 5G} & \tabincell{c}{Google\\Pixel 4\tnote{\dag}} \\ 
        \ChangeRT{0.5pt}
        \hline
        \multirow{13}{*}{\tabincell{c}{UNet-\\ResizeBilinear-\\PRelu}} & None & Float & 31.87 / 0.903 & 256 & \bf{1281} & 1503 & 9770\tnote{1} \\
        \cline{2-8}
        & \multirow{3}{*}{\tabincell{c}{Quantization\\(fixed-point)}} & 8-bit PTQ  & 29.66 / 0.835 & \multirow{3}{*}{256} & 31220\tnote{2} & \bf{504}  & 2175\tnote{3} \\
        &                               & 8-bit QAT  & 31.03 / 0.873 &                      & 33490\tnote{2} & \bf{488}  & 2128\tnote{3}\\
        &                               & 16-bit PTQ\tnote{\ddag} & 31.87 / 0.903 &                      & --             & {\bf 825}\tnote{4}  & -- \\
        \cline{2-8}
        & \multirow{5}{*}{\tabincell{c}{Pruning\\(floating-point)}} & -5\% MAC & 31.85 / 0.903 & 243 & 1693 & \bf{1469} & 8419\tnote{1} \\
        &                          & -10\% MAC & 31.79 / 0.902 & 230 & 1854 & \bf{1416} & 8189\tnote{1} \\
        &                          & -20\% MAC & 31.54 / 0.896 & 202 & 1853 & \bf{1322} & 7313\tnote{1} \\
        &                          & -30\% MAC & 31.35 / 0.893 & 179 & 1690 & \bf{1225} & 3756\tnote{1} \\
        &                          & -50\% MAC & 29.81 / 0.854 & 127 & 1477 & 1068      & \bf{934} \\
        \cline{2-8}
        & \multirow{3}{*}{\tabincell{c}{Pruning + \\Quantization\\(fixed-point)}} & \tabincell{c}{-5\% MAC +\\8-bit QAT} & 31.02 / 0.872 & 243 & 28521\tnote{2} & \bf{482} & 2098\tnote{3} \\
        \cline{3-8}
        &  & \tabincell{c}{-5\% MAC +\\16-bit PTQ} & 31.85 / 0.903 & 243 & -- & {\bf 798}\tnote{4} & --\\
        \ChangeRT{0.5pt}
        \hline
    
    \end{tabular}
    \begin{tablenotes}\scriptsize
    \item PTQ, abbreviation of Post-Training Quantization; QAT, abbreviation of Quantization-Aware Training.
    \item[\dag] In Google Pixel 4, all operations of quantized networks are executed on \emph{qti-default} runtime unless fallbacks on CPU are specified.
    \item[\ddag] For 16-bit fixed-point inference on mobile devices, please refer to Section~\ref{sec:exp_opt} for more details.
    \item[1] CONCATENATION and CONV\_2D after CONCATENATION operations fall back to CPU.
    \item[2] PRELU and RESIZE\_BILINEAR operations fall back to CPU.
    \item[3] CONCATENATION operations fall back to CPU.
    \item[4] Latency evaluated with MediaTek NeuroPilot SDK \cite{neuropilot}.
    
    \end{tablenotes}
    \end{threeparttable}
\end{center}
\end{table*}

\begin{figure*}
\begin{minipage}[t]{0.18\linewidth}
  \centering
  \centerline{\includegraphics[width=3.45cm]
  {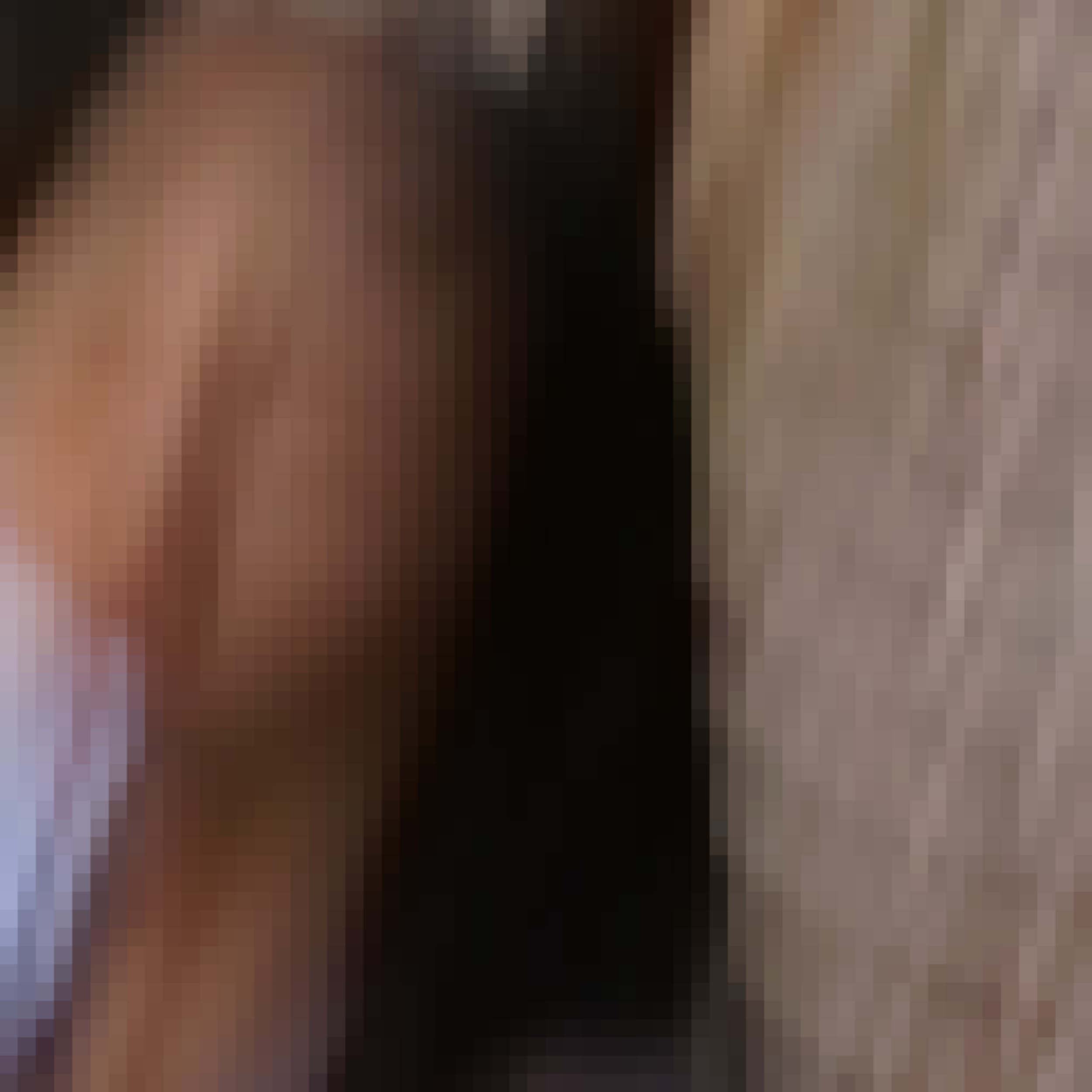}}
  \centerline{(a) Input Patch}    
\end{minipage}
\hspace{0.015\linewidth}
\begin{minipage}[t]{0.18\linewidth}
  \centering
  \centerline{\includegraphics[width=3.45cm]
  {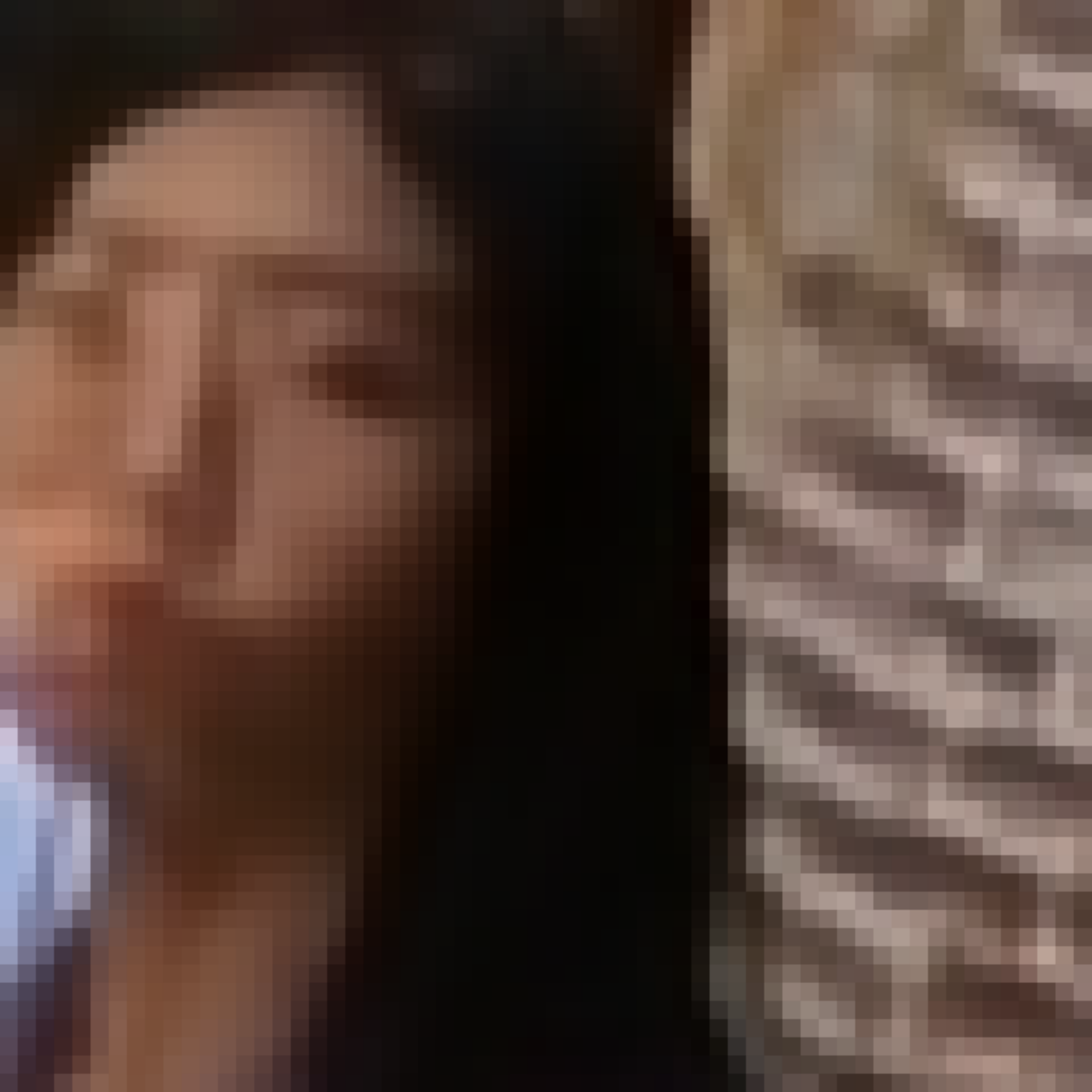}}
  \centerline{\tabincell{c}{(b) Floating-point}}  
\end{minipage}
\hspace{0.015\linewidth}
\begin{minipage}[t]{0.18\linewidth}
  \centering
  \centerline{\includegraphics[width=3.45cm]
  {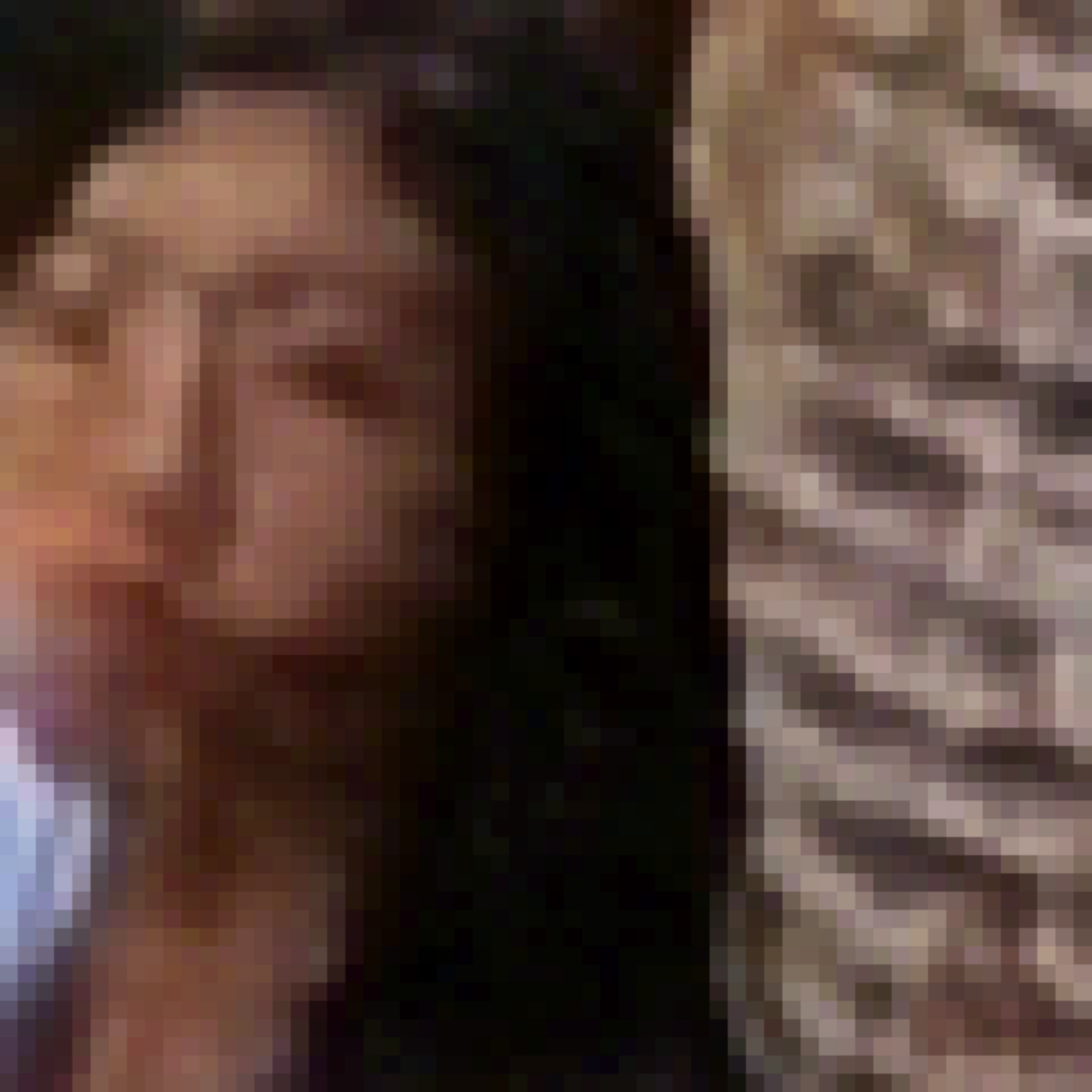}}
  \centerline{\tabincell{c}{(c) 8-bit PTQ}}    
\end{minipage}
\hspace{0.015\linewidth}
\begin{minipage}[t]{0.18\linewidth}
  \centering
  \centerline{\includegraphics[width=3.45cm]
  {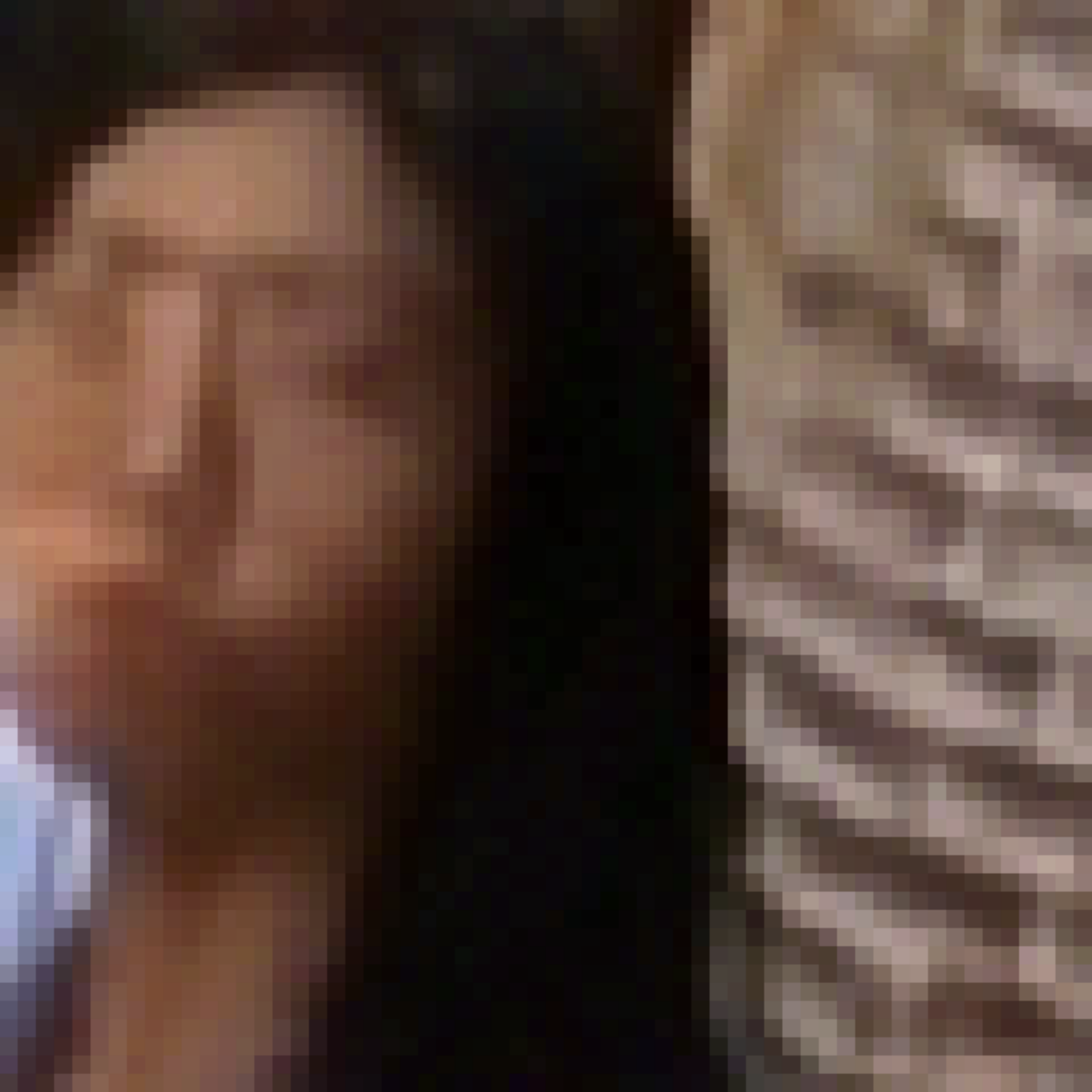}} 
  \centerline{\tabincell{c}{(d) 8-bit QAT}}  
\end{minipage}
\hspace{0.015\linewidth}
\begin{minipage}[t]{0.18\linewidth}
  \centering
  \centerline{\includegraphics[width=3.45cm]
  {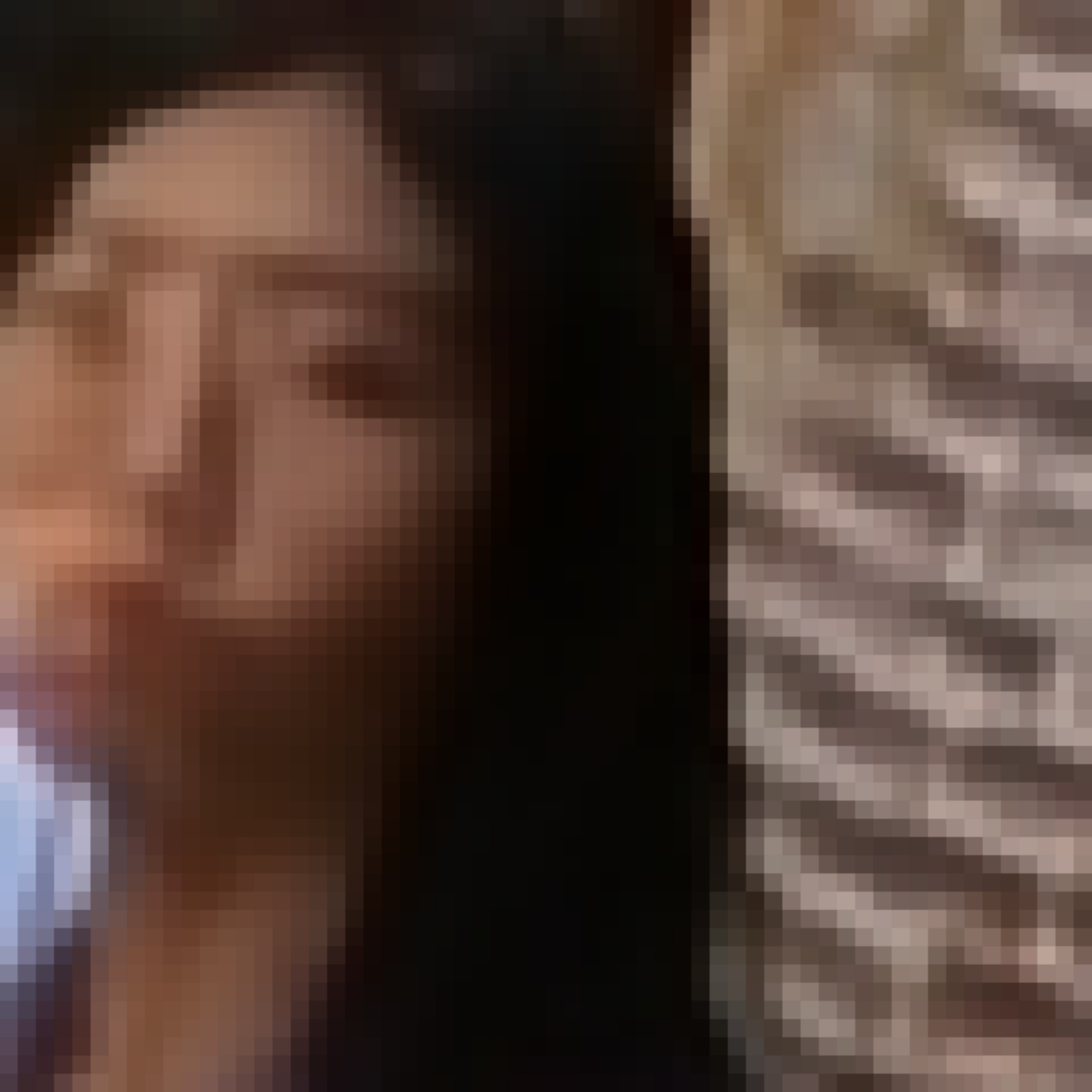}}
  \centerline{\tabincell{c}{(e) 16-bit PTQ}}  
\end{minipage}

\begin{minipage}[t]{0.18\linewidth}
  \centering
  \centerline{\includegraphics[width=3.45cm]
  {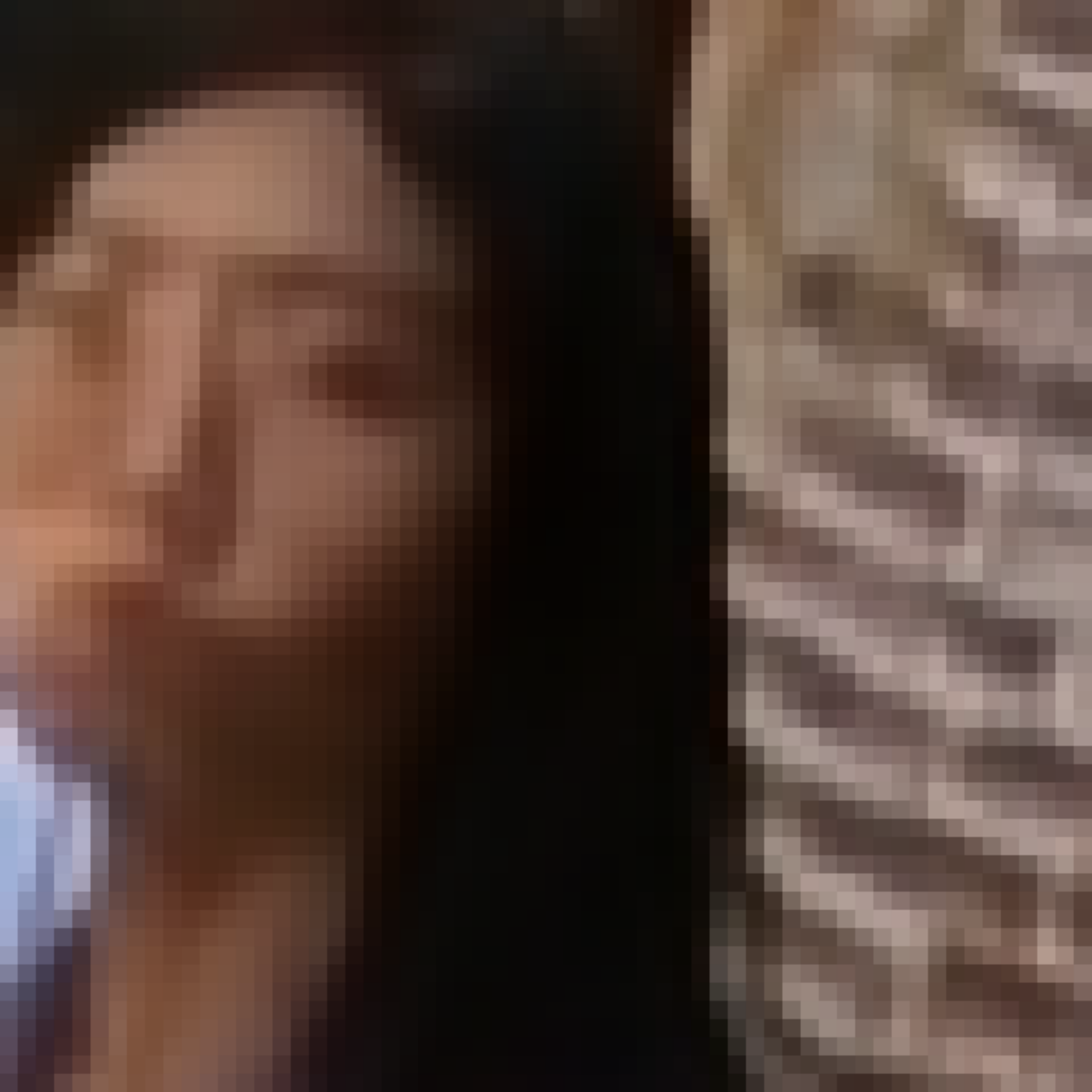}}
  \centerline{\tabincell{c}{(f) -5\% MAC}}  
\end{minipage}
\hspace{0.015\linewidth}
\begin{minipage}[t]{0.18\linewidth}
  \centering
  \centerline{\includegraphics[width=3.45cm]
  {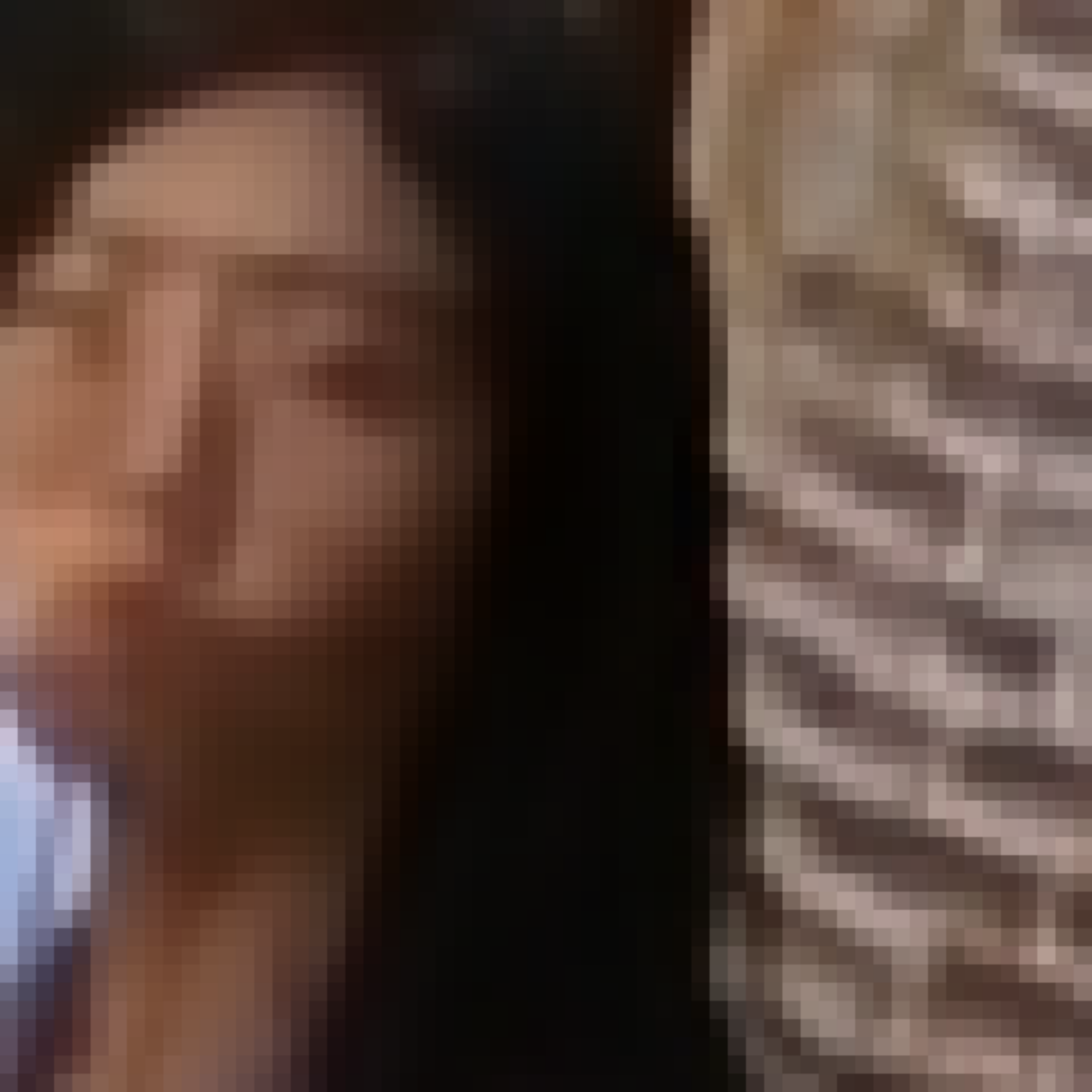}}
  \centerline{\tabincell{c}{(g) -10\% MAC}}  
\end{minipage}
\hspace{0.015\linewidth}
\begin{minipage}[t]{0.18\linewidth}
  \centering
  \centerline{\includegraphics[width=3.45cm]
  {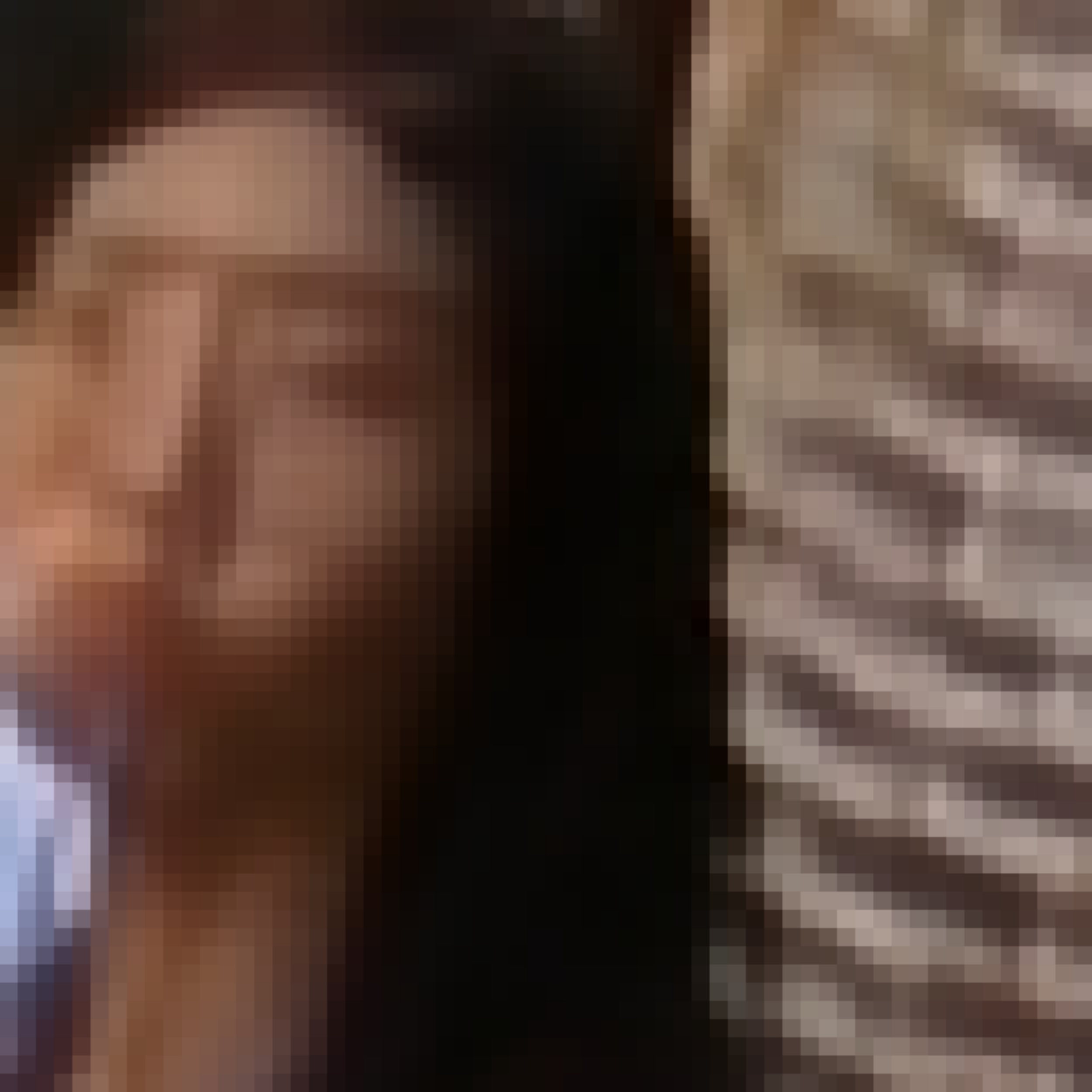}}
  \centerline{\tabincell{c}{(h) -30\% MAC}}
\end{minipage}
\hspace{0.015\linewidth}
\begin{minipage}[t]{0.18\linewidth}
  \centering
  \centerline{\includegraphics[width=3.45cm]
  {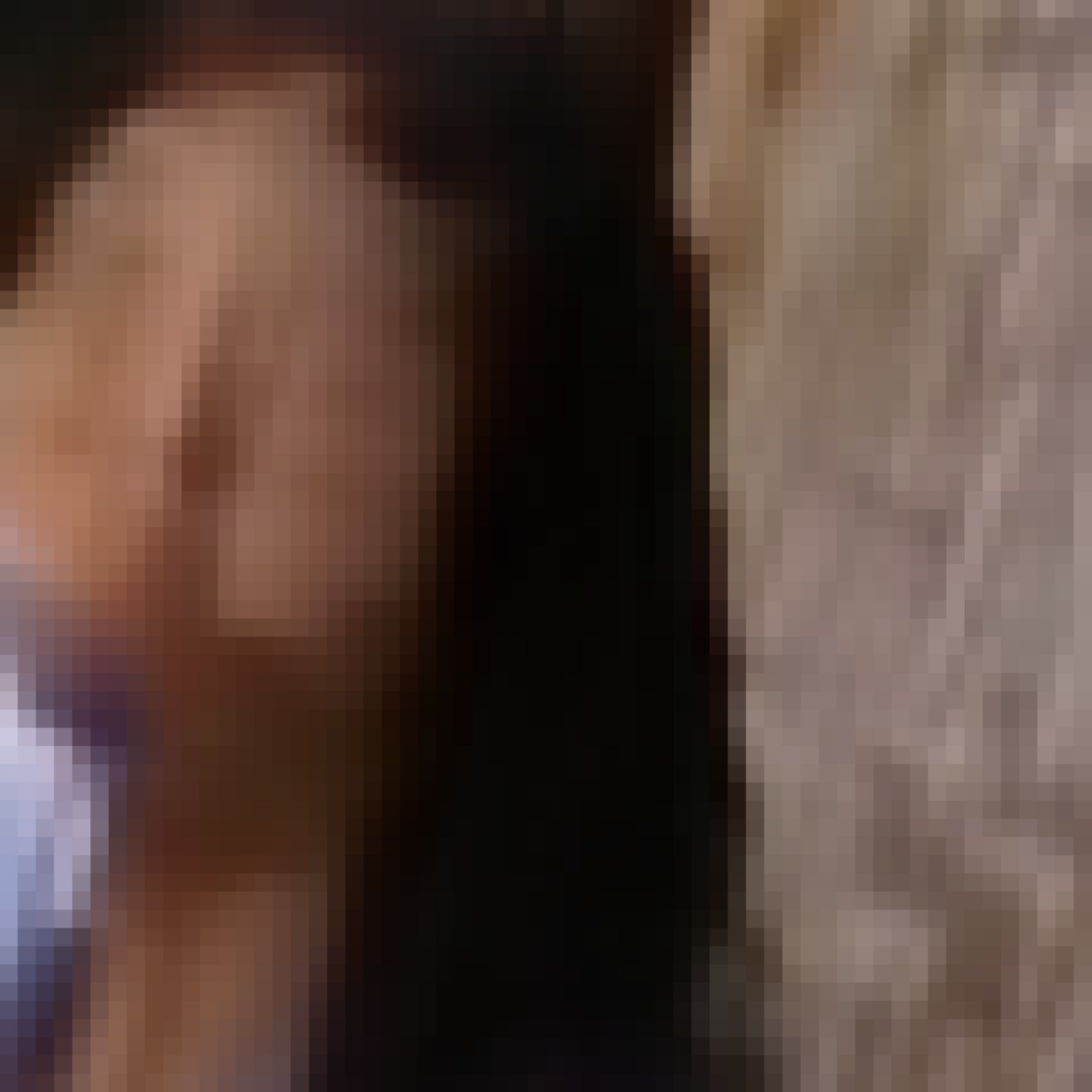}}
  \centerline{\tabincell{c}{(i) -50\% MAC}}
\end{minipage}
\hspace{0.015\linewidth}
\begin{minipage}[t]{0.18\linewidth}
  \centering
  \centerline{\includegraphics[width=3.45cm]
  {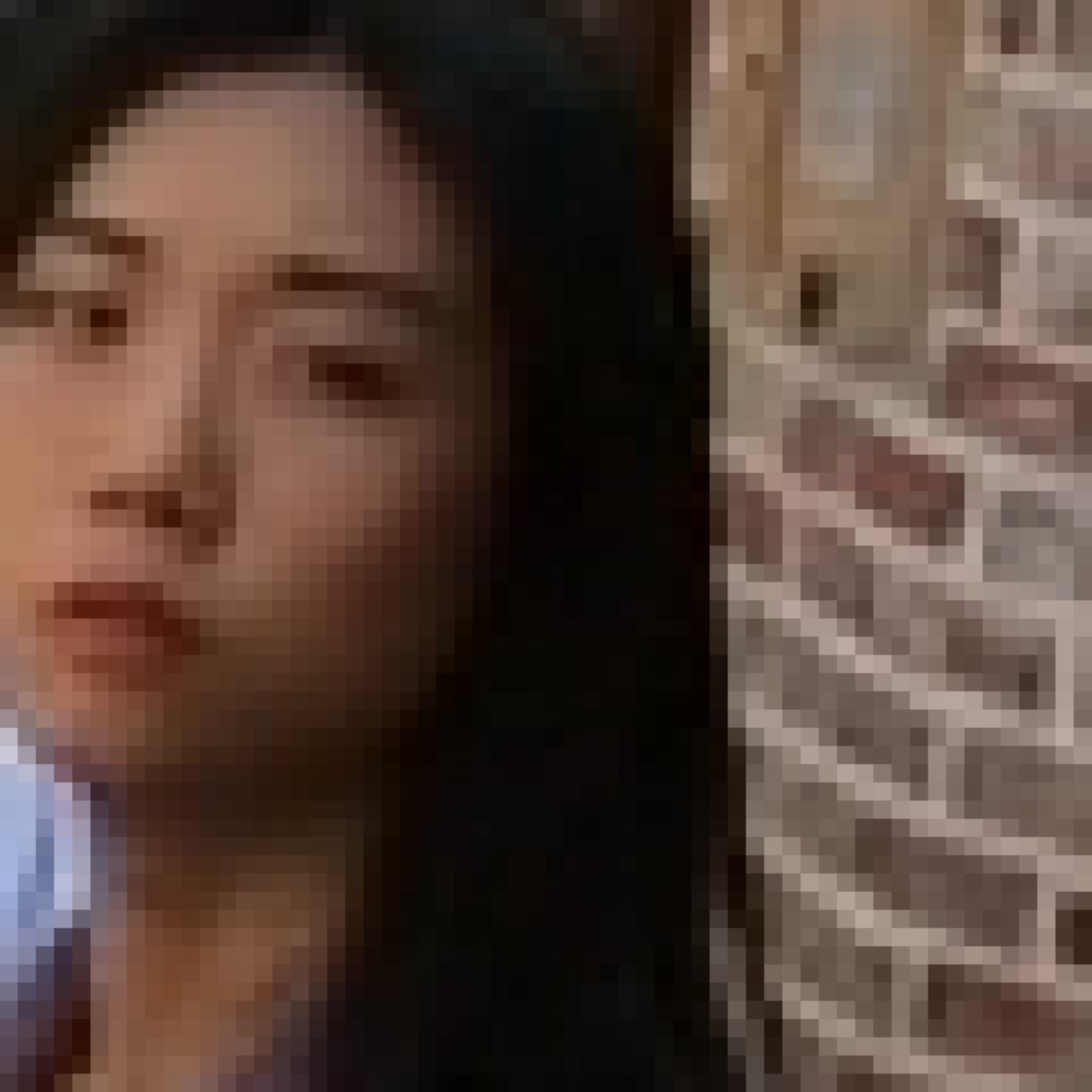}}
  \centerline{(j) Ground Truth Patch}  
\end{minipage}

\caption{Visual results of UNet-ResizeBilinear-PRelu with network quantization and pruning. (c)(d)(e) represent different quantization settings as in Table~\ref{tab:optimizationcompare}. (e)(f)(g)(h) show the results of pruning given different MAC reduction targets. The patch is cropped from \emph{000/00000039.png} in REDS validation set}
\label{fig:visualresult}
\end{figure*}

\begin{table*}
\begin{center}
    \centering
    \caption{Error measurement on mobile devices with various data types. Evaluated on UNet-ResizeBilinear-Relu network\footnotesize{$^\dag$}}.
    \label{tab:quality}
    \begin{threeparttable}[t]
    \begin{tabular}{ c  c  c  c | c  c | c  c}
        \ChangeRT{0.5pt}
        \hline
        \multirow{3}{*}{Data Type} & \multirow{3}{*}{\tabincell{c}{PSNR}} & \multicolumn{6}{c}{Comparison between results on mobile devices (TFLite) and results on Desktop (TensorFlow)} \\
        \cline{3-8} & & \multicolumn{2}{c|}{Huawei Mate30 Pro} & \multicolumn{2}{c|}{OPPO Reno3 5G} & \multicolumn{2}{c}{Google Pixel 4} \\ 
        & & PSNR & L2 Error & PSNR & L2 Error & PSNR & L2 Error \\
        \hline\hline
        Float\tnote{1}  & 31.65 & 50.82 $\pm$ 0.29 & 1.95$\times10^{-7}$ & {\bf 50.85} $\pm$ \bf{0.19} & \bf{1.94}$\times\bf{10^{-7}}$ & 50.59 $\pm$ 0.73 & 2.09$\times10^{-7}$ \\
        16-bit & 31.65 & --               & --                  & {\bf 65.37 $\pm$ 0.89}       & {\bf 6.99}$\times\bf{10^{-9}}$ & -- & --\\
        8-bit  & 31.36 & 43.07 $\pm$ 1.52 & 1.25$\times10^{-6}$ & {\bf 43.33} $\pm$ \bf{1.39} & \bf{1.16}$\times\bf{10^{-6}}$ & 41.95 $\pm$ 1.24 & 1.57$\times10^{-6}$ \\
        \ChangeRT{0.5pt}
        \hline

    \end{tabular}
    \begin{tablenotes}\scriptsize
    \item Standard deviation of PSNR is calculated with 300 validation images.
    \item[\dag] Abnormal PSNR drops (for Floating-point setting) happen to UNet-ResizeBilinear-PRelu network on Google Pixel 4. Since the root cause is not confirmed, this table reports the results of UNet-ResizeBilinear-Relu network for a fair comparison.
    \item[1] Floating-point data type used for mobile and desktop are 16-bit and 32-bit, respectively
    \end{tablenotes}
    \end{threeparttable}
\end{center}
\end{table*}

\vspace{10pt}
\subsection{Network Optimization}
\label{sec:exp_opt}
According to the discussion in Section~\ref{sec:exp_ops}, this paper selects \emph{UNet-ResizeBilinear-PRelu} and applies network optimizations to further boost its performance.
Experiment results of network optimization are summarized in Table~\ref{tab:optimizationcompare}.

\vspace{-5pt}
\subsubsection{Quantization.}
For 8-bit quantization, post-training quantization suffers a destructive 2 dB PSNR drop.
Even with quantization-aware training, there exists at least noticeable 0.8 dB PSNR drop.
In contrast, 16-bit post-training quantization, is capable to preserve almost the same quality as floating-point network. 
In our experiments, most devices have latency improvement with quantized networks except for Huawei Mate30 Pro.
This is due to the lack of support for quantized \emph{RESIZE\_BILINEAR} operation in its accelerator.
We suggest future works to consider quantization configuration during the stage of architecture search.
Note that NNAPI does not support 16-bit fixed-point inference. Hence, the evaluation requires proprietary SDK provided by platform providers. Qualcomm's SNPE \cite{snpe} supports 16-bit fixed-point inference with HTA hardware. However, the corresponding software (HTA runtime library) is not available in Google Pixel 4. For Huawei's HiAI SDK \cite{hiai}, we cannot find appropriate information for its support of 16-bit fixed-point inference. Therefore, we only report the results of 16-bit fixed-point inference for MediaTek's NeuroPilot SDK \cite{neuropilot} in Table~\ref{tab:optimizationcompare}.

\vspace{-5pt}
\subsubsection{Pruning.}
We apply five different settings of MAC reduction targets. 
Most devices have latency improvement except Huawei Mate30 Pro. 
Surprisingly, we observe over 60\% latency improvement with roughly 0.5 db PSNR drop on Google Pixel 4 (in 30\% MAC reduction setting). 
Hence, as a future direction, such hardware limitations and preferences should also be considered when searching network architectures.

Last, we combine both network pruning and quantization for further optimization.
Based on the network pruned with 5\% MAC reduction, we quantize the network with 8-bit quantization-aware training and 16-bit post-training quantization.
As shown in Table~\ref{tab:optimizationcompare}, the latency could be further reduced when compared with quantization only.

\begin{figure}[t]
\begin{minipage}[b]{1\linewidth}
  \centering
  \centerline{\includegraphics[width=8.4cm]
  {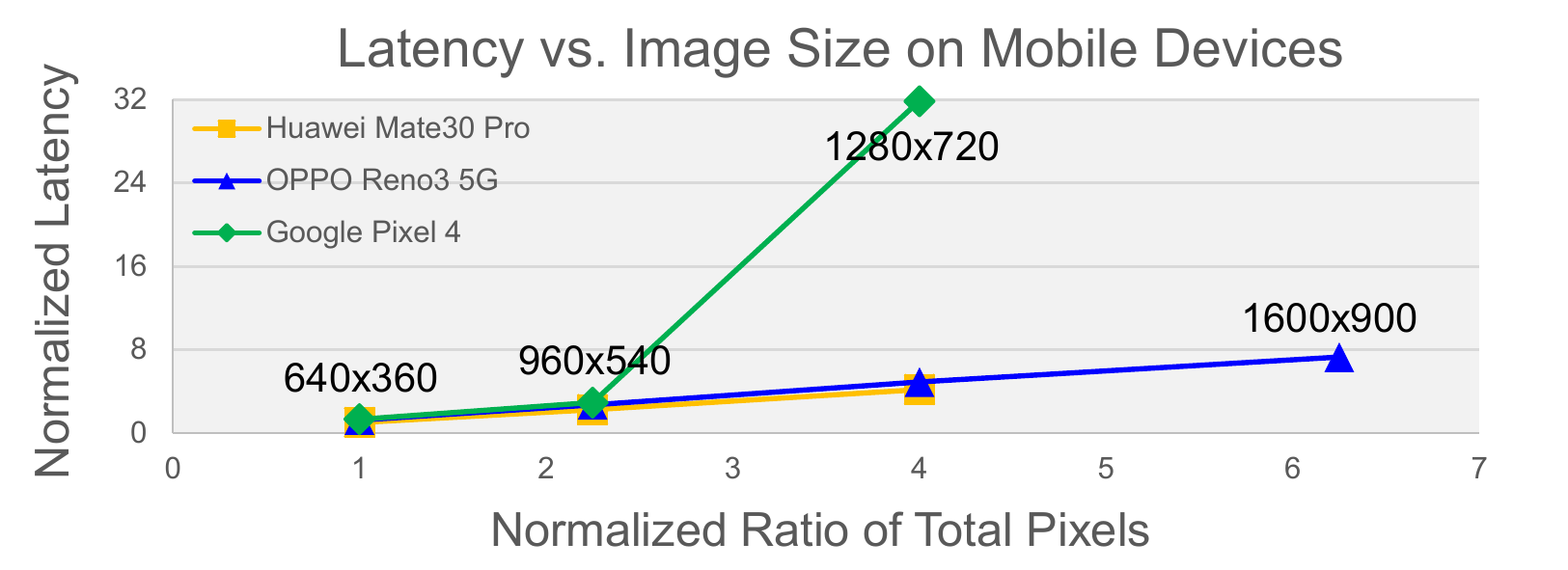}}
  \medskip
\end{minipage}
\vspace{-15pt}
\caption{Latency with various input resolution. 
All the latencies are normalized to Huawei Mate30 Pro with 360p resolution. Note that only OPPO Reno3 5G successfully run with 1600$\times$900 (HD+) input resolution. All devices fail to run with 1920$\times$1080 (Full HD resolution. }
\label{fig:latencyres}
\end{figure}

\vspace{10pt}
\subsection{Ablation Study of Quality}
In this section, we show the impact of different network optimization by examining visual results.
The quantitative results are also illustrated to uncover the computation errors across different hardware implementations.
\label{sec:exp_quality}

\vspace{-5pt}
\subsubsection{Visual Quality on Optimized Networks}
Figure~\ref{fig:visualresult} shows visual results of quantization and pruning.
16-bit post-training quantization (PTQ) perfectly preserve the visual quality of floating-point network.
However, 8-bit post-training quantization (PTQ) and quantization-aware training (QAT) show different levels of quantization errors.
For pruning results, the visual quality degrades with the increasing of MAC reduction. 
When pruning the network by 50\% MAC, a noticeable blurry result appears. 

\subsubsection{Quality Index on Mobile Devices}
Table~\ref{tab:quality} shows the PSNR and per-pixel L2 error.
Such calculations are between the results of TensorFlow (\emph{checkpoint} format) on desktops and the results of TFLite on mobile devices.
In floating-point data type, 32-bit data are used in TensorFlow inference while mobile devices use 16-bit data for acceleration.
PSNR and L2 error between 32-bit and 16-bit floating-point are about 50 dB and 1.94$\times10^{-7}$, respectively.
For quantized data type, TensorFlow uses \emph{fake quantitzation} operations \cite{tfquant} to simulate the behavior of quantization.
However, the inference is still computed by using floating-point arithmetic, which is different from the fixed-point ones used by TFLite.
As shown in the table, the error of 8-bit data type is much larger than 16-bit and floating-point data type.
This provides an in-depth quality assessment for deploying quantized networks on mobile devices. 
In general, the lower error between mobile devices and desktops, the closer result between algorithm development and its deployment on devices.

\subsection{Discussions}

Considering the differences of software and hardware between all the three platforms, several non-trial deployment issues are reported in the previous sections.
This section summarizes all the findings and discussions.

\begin{itemize}
\item {\bf First of all,}
as listed in Table \ref{tab:archicompare}, the latency are highly inconsistent when deploying the out-of-the-box network architectures across platforms. 
Some of platforms (Huawei Mate30 Pro and Google Pixel 4) present unreasonably high latency. Fortunately, as in Table \ref{tab:opcompare}, such  pitfall can be partially mitigated by leveraging operations with better portability.
\item {\bf Second,}
the network optimization techniques (both quantization and pruning) do not consistently reduce the latency across platforms.
As listed in Table \ref{tab:optimizationcompare}, network optimizations cause even higher latency in Huawei Mate30 Pro.
Meanwhile, the latency of pruned networks do not scale linearly w.r.t. MAC reduction in Google Pixel 4.
\item {\bf Last but not the least,}
as shown in Figure~\ref{fig:latencyres}, the latency does not scale linearly with input resolution. 
In Google Pixel 4, there is a huge latency increment when the input resolution scales to $1280\times720$.
\end{itemize}

In summary, these non-trivial performance pitfalls make mobile deployment an even challenging work. This urges deployment-guidelines to conduct 1) portable network architectures, 2) network optimization and 3) trade between quality and latency across mobile devices.

\section{Conclusion and Future Work}
\label{sec:conclusion}
In summary, this paper conducts a search of portable network architectures for better quality-latency trade-off across mobile devices.
Besides, we also present the effectiveness of quantization and pruning for image deblurring task.
The searched portable networks are evaluated with a set of comprehensive experiments and comparisons.
Our experiments and comparisons provide an in-depth analysis for both latency and image quality.
In conclusion, we demonstrate a success deployment of image deblurring on three mobile devices.
We also suggest two promising directions for future works (1) searching portable network architecture while considering more device related factors, \eg, quantization, pruning and/or hardware limitation/preference, and (2) systematic searching methodology for portable network architecture, \eg, Network Architecture Search (NAS) for device portability.

{
\small
\bibliographystyle{ieee_fullname}

\begin{thebibliography}{10}

\bibitem{aibenchmarkweb}
{AI Benchmark Performance Ranking}.
\newblock \url{http://ai-benchmark.com/ranking.html}.

\bibitem{ludashi}
{AImark of Ludashi}.
\newblock \url{http://www.ludashi.com/page/aimark.php}.

\bibitem{nnapi}
{Android Neural Networkk API (NNAPI)}.
\newblock \url{https://developer.android.com/ndk/guides/neuralnetworks}.

\bibitem{antutu}
{Antutu AI Benchmark},.

\bibitem{iphone11}
{Apple A13 Bionic Chipset}.
\newblock \url{https://en.wikichip.org/wiki/apple/ax/a13}.

\bibitem{phonecompare}
{Comparison of Huawei Mate30 Pro 5G, OPPO Reno3 and Google Pixel 4}.
\newblock
  \url{https://www.gsmarena.com/compare.php3?&idPhone1=9880&idPhone2=9942&idPhone3=9896}.

\bibitem{razarp20}
{Comparison of Razer Phone and Huawei P20}.
\newblock
  \url{https://www.gsmarena.com/compare.php3?idPhone1=8923&idPhone2=9107}.

\bibitem{hiai}
{Huawei HiAI SDK}.
\newblock \url{https://developer.huawei.com/consumer/en/hiai}.

\bibitem{kirin990}
{Huawei Kirin 990 5G Chipset}.
\newblock \url{https://en.wikichip.org/wiki/Kirin_990}.

\bibitem{dimensity1000}
{MediaTek Dimensity 1000L 5G Chipset}.
\newblock
  \url{https://en.wikichip.org/wiki/mediatek/dimensity/1000l#Neural_processor}.

\bibitem{neuropilot}
{MediaTek NeuroPilot SDK}.
\newblock \url{https://neuropilot.mediatek.com/}.

\bibitem{s855}
{Qualcomm Snapdragon 855 Chipset}.
\newblock \url{https://en.wikichip.org/wiki/qualcomm/snapdragon_800/855}.

\bibitem{snpe}
{Qualcomm Snapdragon Neural Processing Engine SDK}.
\newblock \url{https://developer.qualcomm.com/docs/snpe/overview.html}.

\bibitem{tfdeeplabquant}
{Quantize DeepLab Model for Faster on-device Inference}.
\newblock
  \url{https://github.com/tensorflow/models/blob/master/research/deeplab/g3doc/quantize.md}.

\bibitem{exynos990}
{Samsung Exynos 990 Mobile Processor}.
\newblock \url{https://en.wikichip.org/wiki/samsung/exynos/990}.

\bibitem{tfdetection}
{Tensorflow Detection Model Zoo}.
\newblock
  \url{https://github.com/tensorflow/models/blob/master/research/object_detection/g3doc/detection_model_zoo.md}.

\bibitem{benchmarktool}
{TFLite Model Benchmark Tool}.
\newblock
  \url{https://github.com/tensorflow/tensorflow/tree/master/tensorflow/lite/tools/benchmark}.

\bibitem{carn}
Namhyuk Ahn, Byungkon Kang, and Kyung-Ah Sohn.
\newblock Fast, accurate, and lightweight super-resolution with cascading
  residual network.
\newblock In {\em Proceedings of the European Conference on Computer Vision
  (ECCV)}, pages 252--268, 2018.

\bibitem{sid}
Chen Chen, Qifeng Chen, Jia Xu, and Vladlen Koltun.
\newblock Learning to see in the dark.
\newblock In {\em Proceedings of the IEEE Conference on Computer Vision and
  Pattern Recognition}, pages 3291--3300, 2018.

\bibitem{srquant}
Yoojin Choi, Mostafa El-Khamy, and Jungwon Lee.
\newblock Learning low precision deep neural networks through regularization.
\newblock {\em arXiv preprint arXiv:1809.00095}, 2018.

\bibitem{falsr}
Xiangxiang Chu, Bo Zhang, Hailong Ma, Ruijun Xu, Jixiang Li, and Qingyuan Li.
\newblock Fast, accurate and lightweight super-resolution with neural
  architecture search.
\newblock {\em arXiv preprint arXiv:1901.07261}, 2019.

\bibitem{dynamicscenedeblur}
Hongyun Gao, Xin Tao, Xiaoyong Shen, and Jiaya Jia.
\newblock Dynamic scene deblurring with parameter selective sharing and nested
  skip connections.
\newblock In {\em Proceedings of the IEEE Conference on Computer Vision and
  Pattern Recognition}, pages 3848--3856, 2019.

\bibitem{xavier}
Xavier Glorot and Yoshua Bengio.
\newblock Understanding the difficulty of training deep feedforward neural
  networks.
\newblock In {\em Proceedings of the thirteenth international conference on
  artificial intelligence and statistics}, pages 249--256, 2010.

\bibitem{morphnet}
Ariel Gordon, Elad Eban, Ofir Nachum, Bo Chen, Hao Wu, Tien-Ju Yang, and Edward
  Choi.
\newblock Morphnet: Fast \& simple resource-constrained structure learning of
  deep networks.
\newblock In {\em Proceedings of the IEEE Conference on Computer Vision and
  Pattern Recognition}, pages 1586--1595, 2018.

\bibitem{mtlu}
Shuhang Gu, Wen Li, Luc~Van Gool, and Radu Timofte.
\newblock Fast image restoration with multi-bin trainable linear units.
\newblock In {\em Proceedings of the IEEE International Conference on Computer
  Vision}, pages 4190--4199, 2019.

\bibitem{sgn}
Shuhang Gu, Yawei Li, Luc~Van Gool, and Radu Timofte.
\newblock Self-guided network for fast image denoising.
\newblock In {\em Proceedings of the IEEE International Conference on Computer
  Vision}, pages 2511--2520, 2019.

\bibitem{deepcompression}
Song Han, Huizi Mao, and William~J Dally.
\newblock Deep compression: Compressing deep neural networks with pruning,
  trained quantization and huffman coding.
\newblock {\em arXiv preprint arXiv:1510.00149}, 2015.

\bibitem{dbpn}
Muhammad Haris, Gregory Shakhnarovich, and Norimichi Ukita.
\newblock Deep back-projection networks for super-resolution.
\newblock In {\em Proceedings of the IEEE conference on computer vision and
  pattern recognition}, pages 1664--1673, 2018.

\bibitem{resnet}
Kaiming He, Xiangyu Zhang, Shaoqing Ren, and Jian Sun.
\newblock Deep residual learning for image recognition.
\newblock In {\em Proceedings of the IEEE conference on computer vision and
  pattern recognition}, pages 770--778, 2016.

\bibitem{mobilenet}
Andrew~G Howard, Menglong Zhu, Bo Chen, Dmitry Kalenichenko, Weijun Wang,
  Tobias Weyand, Marco Andreetto, and Hartwig Adam.
\newblock Mobilenets: Efficient convolutional neural networks for mobile vision
  applications.
\newblock {\em arXiv preprint arXiv:1704.04861}, 2017.

\bibitem{idn}
Zheng Hui, Xiumei Wang, and Xinbo Gao.
\newblock Fast and accurate single image super-resolution via information
  distillation network.
\newblock In {\em Proceedings of the IEEE conference on computer vision and
  pattern recognition}, pages 723--731, 2018.

\bibitem{aibenchmark}
Andrey Ignatov, Radu Timofte, Andrei Kulik, Seungsoo Yang, Ke Wang, Felix Baum,
  Max Wu, Lirong Xu, and Luc Van~Gool.
\newblock Ai benchmark: All about deep learning on smartphones in 2019.
\newblock {\em arXiv preprint arXiv:1910.06663}, 2019.

\bibitem{pirmsr}
Andrey Ignatov, Radu Timofte, Thang Van~Vu, Tung Minh~Luu, Trung X~Pham, Cao
  Van~Nguyen, Yongwoo Kim, Jae-Seok Choi, Munchurl Kim, Jie Huang, et~al.
\newblock Pirm challenge on perceptual image enhancement on smartphones:
  Report.
\newblock In {\em Proceedings of the European Conference on Computer Vision
  (ECCV)}, pages 0--0, 2018.

\bibitem{pynet}
Andrey {Ignatov}, Luc {Van Gool}, and Radu {Timofte}.
\newblock {Replacing Mobile Camera ISP with a Single Deep Learning Model}.
\newblock {\em arXiv e-prints}, page arXiv:2002.05509, Feb 2020.

\bibitem{tfquant}
Benoit Jacob, Skirmantas Kligys, Bo Chen, Menglong Zhu, Matthew Tang, Andrew
  Howard, Hartwig Adam, and Dmitry Kalenichenko.
\newblock Quantization and training of neural networks for efficient
  integer-arithmetic-only inference.
\newblock In {\em Proceedings of the IEEE Conference on Computer Vision and
  Pattern Recognition}, pages 2704--2713, 2018.

\bibitem{adam}
Diederik~P Kingma and Jimmy Ba.
\newblock Adam: A method for stochastic optimization.
\newblock {\em arXiv preprint arXiv:1412.6980}, 2014.

\bibitem{deblurganv2}
Orest Kupyn, Tetiana Martyniuk, Junru Wu, and Zhangyang Wang.
\newblock Deblurgan-v2: Deblurring (orders-of-magnitude) faster and better.
\newblock In {\em Proceedings of the IEEE International Conference on Computer
  Vision}, pages 8878--8887, 2019.

\bibitem{srgan}
Christian Ledig, Lucas Theis, Ferenc Husz{\'a}r, Jose Caballero, Andrew
  Cunningham, Alejandro Acosta, Andrew Aitken, Alykhan Tejani, Johannes Totz,
  Zehan Wang, et~al.
\newblock Photo-realistic single image super-resolution using a generative
  adversarial network.
\newblock In {\em Proceedings of the IEEE conference on computer vision and
  pattern recognition}, pages 4681--4690, 2017.

\bibitem{noise2noise}
Jaakko Lehtinen, Jacob Munkberg, Jon Hasselgren, Samuli Laine, Tero Karras,
  Miika Aittala, and Timo Aila.
\newblock Noise2noise: Learning image restoration without clean data.
\newblock {\em arXiv preprint arXiv:1803.04189}, 2018.

\bibitem{oicsr}
Jiashi Li, Qi Qi, Jingyu Wang, Ce Ge, Yujian Li, Zhangzhang Yue, and Haifeng
  Sun.
\newblock Oicsr: Out-in-channel sparsity regularization for compact deep neural
  networks.
\newblock In {\em Proceedings of the IEEE Conference on Computer Vision and
  Pattern Recognition}, pages 7046--7055, 2019.

\bibitem{edsr}
Bee Lim, Sanghyun Son, Heewon Kim, Seungjun Nah, and Kyoung Mu~Lee.
\newblock Enhanced deep residual networks for single image super-resolution.
\newblock In {\em Proceedings of the IEEE conference on computer vision and
  pattern recognition workshops}, pages 136--144, 2017.

\bibitem{fpn}
Tsung-Yi Lin, Piotr Doll{\'a}r, Ross Girshick, Kaiming He, Bharath Hariharan,
  and Serge Belongie.
\newblock Feature pyramid networks for object detection.
\newblock In {\em Proceedings of the IEEE conference on computer vision and
  pattern recognition}, pages 2117--2125, 2017.

\bibitem{megviinr}
Jiaming Liu, Chi-Hao Wu, Yuzhi Wang, Qin Xu, Yuqian Zhou, Haibin Huang, Chuan
  Wang, Shaofan Cai, Yifan Ding, Haoqiang Fan, et~al.
\newblock Learning raw image denoising with bayer pattern unification and bayer
  preserving augmentation.
\newblock In {\em Proceedings of the IEEE Conference on Computer Vision and
  Pattern Recognition Workshops}, pages 0--0, 2019.

\bibitem{binarysr}
Yinglan Ma, Hongyu Xiong, Zhe Hu, and Lizhuang Ma.
\newblock Efficient super resolution using binarized neural network.
\newblock In {\em Proceedings of the IEEE Conference on Computer Vision and
  Pattern Recognition Workshops}, pages 0--0, 2019.

\bibitem{reds}
Seungjun Nah, Sungyong Baik, Seokil Hong, Gyeongsik Moon, Sanghyun Son, Radu
  Timofte, and Kyoung~Mu Lee.
\newblock Ntire 2019 challenges on video deblurring and super-resolution:
  Dataset and study.
\newblock In {\em The IEEE Conference on Computer Vision and Pattern
  Recognition (CVPR) Workshops}, June 2019.

\bibitem{ntiredeblur}
Seungjun Nah, Sanghyun Son, Radu Timofte, and Kyoung~Mu Lee.
\newblock Ntire 2020 challenge on image and video deblurring.
\newblock In {\em The IEEE Conference on Computer Vision and Pattern
  Recognition (CVPR) Workshops}, June 2020.

\bibitem{deformabledeblur}
Kuldeep {Purohit} and A.~N. {Rajagopalan}.
\newblock {Region-Adaptive Dense Network for Efficient Motion Deblurring}.
\newblock {\em arXiv e-prints}, page arXiv:1903.11394, Mar. 2019.

\bibitem{aliveblur}
Kuldeep Purohit, Anshul Shah, and AN Rajagopalan.
\newblock Bringing alive blurred moments.
\newblock In {\em Proceedings of the IEEE Conference on Computer Vision and
  Pattern Recognition}, pages 6830--6839, 2019.

\bibitem{mlperf}
Vijay~Janapa Reddi, Christine Cheng, David Kanter, Peter Mattson, Guenther
  Schmuelling, Carole-Jean Wu, Brian Anderson, Maximilien Breughe, Mark
  Charlebois, William Chou, et~al.
\newblock Mlperf inference benchmark.
\newblock {\em arXiv preprint arXiv:1911.02549}, 2019.

\bibitem{unet}
Olaf Ronneberger, Philipp Fischer, and Thomas Brox.
\newblock U-net: Convolutional networks for biomedical image segmentation.
\newblock In {\em International Conference on Medical image computing and
  computer-assisted intervention}, pages 234--241. Springer, 2015.

\bibitem{inceptionv3}
Christian Szegedy, Vincent Vanhoucke, Sergey Ioffe, Jon Shlens, and Zbigniew
  Wojna.
\newblock Rethinking the inception architecture for computer vision.
\newblock In {\em Proceedings of the IEEE conference on computer vision and
  pattern recognition}, pages 2818--2826, 2016.

\bibitem{mobiledeblur}
Jun Tan, Kang Yang, Shiwei Song, Tianzhang Xing, and Dingyi Fang.
\newblock Mobile-deblur: A clear image will on the smart device.
\newblock In {\em 2017 3rd International Conference on Big Data Computing and
  Communications (BIGCOM)}, pages 97--105. IEEE, 2017.

\bibitem{scalerecurrent}
Xin Tao, Hongyun Gao, Xiaoyong Shen, Jue Wang, and Jiaya Jia.
\newblock Scale-recurrent network for deep image deblurring.
\newblock In {\em Proceedings of the IEEE Conference on Computer Vision and
  Pattern Recognition}, pages 8174--8182, 2018.

\bibitem{mtkprune}
Wei-Ting Wang, Han-Lin Li, Wei-Shiang Lin, Cheng-Ming Chiang, and Yi-Min Tsai.
\newblock Architecture-aware network pruning for vision quality applications.
\newblock In {\em 2019 IEEE International Conference on Image Processing
  (ICIP)}, pages 2701--2705. IEEE, 2019.

\bibitem{mitprune}
Tien-Ju Yang, Yu-Hsin Chen, and Vivienne Sze.
\newblock Designing energy-efficient convolutional neural networks using
  energy-aware pruning.
\newblock In {\em Proceedings of the IEEE Conference on Computer Vision and
  Pattern Recognition}, pages 5687--5695, 2017.

\bibitem{gatedecorator}
Zhonghui You, Kun Yan, Jinmian Ye, Meng Ma, and Ping Wang.
\newblock Gate decorator: Global filter pruning method for accelerating deep
  convolutional neural networks.
\newblock In {\em Advances in Neural Information Processing Systems}, pages
  2130--2141, 2019.

\bibitem{rdn}
Yulun Zhang, Yapeng Tian, Yu Kong, Bineng Zhong, and Yun Fu.
\newblock Residual dense network for image super-resolution.
\newblock In {\em Proceedings of the IEEE conference on computer vision and
  pattern recognition}, pages 2472--2481, 2018.

\end{thebibliography}
\let\oldbibliography\thebibliography
\renewcommand{\thebibliography}[1]{%
  \oldbibliography{#1}%
  \setlength{\itemsep}{0pt}%
}

\end{document}